%% file: openEA.tex
\documentclass{vldb}

\usepackage{adjustbox}
\usepackage[ruled,vlined,linesnumbered]{algorithm2e}
\usepackage{amsfonts}
\usepackage{amssymb}
\usepackage{balance} 
\usepackage{bm}
\usepackage{colortbl}
\usepackage{graphicx}
\usepackage{hyperref}
\usepackage{mathtools}
\usepackage{microtype}
\usepackage{multirow}
\usepackage{scrextend}
\usepackage{soul}
\usepackage{subfig}
\usepackage{xcolor}
\usepackage{xstring}
\usepackage{balance}

\usepackage{caption}
\usepackage[labelfont=bf]{caption}

\newenvironment{noindlist}{
\begin{list}{\labelitemi}{
\leftmargin=1.0em 
}}{\end{list}}

\vldbTitle{A Benchmarking Study of Embedding-based Entity Alignment for Knowledge Graphs}
\vldbAuthors{Zequn Sun, Qingheng Zhang, Wei Hu, Chengming Wang, Muhao Chen, Farahnaz Akrami, Chengkai Li}
\vldbDOI{https://doi.org/10.14778/3407790.3407828}
\vldbVolume{13}
\vldbNumber{11}
\vldbYear{2020}

\begin{document}


 \makeatletter
 \def\@copyrightspace{\relax}
 \makeatother

 \def\endabstract{\if@twocolumn\else\endquotation\fi}


\title{A Benchmarking Study of\\ Embedding-based Entity Alignment for Knowledge Graphs}



%
%
%
%

\numberofauthors{1} 

\author{
%
\alignauthor Zequn Sun\,\raisebox{4pt}{$\dag$}, Qingheng Zhang\,\raisebox{4pt}{$\dag$}, Wei Hu\,\raisebox{4pt}{$\dag$}\titlenote{Wei Hu is the corresponding author.}, 
                  Chengming Wang\,\raisebox{4pt}{$\dag$}, \\
                  Muhao Chen\,\raisebox{4pt}{$\ddag$}, Farahnaz Akrami\,\raisebox{4pt}{$\S$}, Chengkai Li\,\raisebox{4pt}{$\S$}\\
\raisebox{4pt}{$\dag$}\,\affaddr{State Key Laboratory for Novel Software Technology, Nanjing University, China}\\
\raisebox{4pt}{$\ddag$}\,\affaddr{Department of Computer Science, University of California, Los Angeles, USA}\\
\raisebox{4pt}{$\S$}\,\affaddr{Department of Computer Science and Engineering, University of Texas at Arlington, USA}\\
\email{
\{zqsun, qhzhang, cmwang\}.nju@gmail.com, whu@nju.edu.cn, muhaochen@ucla.edu, farahnaz.akrami@mavs.uta.edu, cli@uta.edu}
}

\maketitle
\begin{abstract}
Entity alignment seeks to find entities in different knowledge graphs (KGs) that refer to the same real-world object. Recent advancement in KG embedding impels the advent of embedding-based entity alignment, which encodes entities in a continuous embedding space and measures entity similarities based on the learned embeddings. In this paper, we conduct a comprehensive experimental study of this emerging field. We survey 23 recent embedding-based entity alignment approaches and categorize them based on their techniques and characteristics. We also propose a new KG sampling algorithm, with which we generate a set of dedicated benchmark datasets with various heterogeneity and distributions for a realistic evaluation. We develop an open-source library including 12 representative embedding-based entity alignment approaches, and extensively evaluate these approaches, to understand their strengths and limitations. Additionally, for several directions that have not been explored in current approaches, we perform exploratory experiments and report our preliminary findings for future studies. The benchmark datasets, open-source library and experimental results are all accessible online and will be duly maintained. 
\end{abstract}

\section{Introduction}
\label{sect:intro}

Knowledge graphs (KGs) store facts as triples in the form of (\textit{subject entity, relation, object entity}) or (\textit{subject entity, attribute, literal value}). This type of knowledge bases supports a variety of applications, e.g., semantic search, question answering and recommender systems \cite{KDDTutorial}. To promote knowledge fusion, researchers have made considerable progress on the task of \emph{entity alignment}, which is also often termed \emph{entity matching} or \emph{entity resolution}. The goal is to identify entities from different KGs that refer to the same entity, e.g., \textsf{Mount\_Everest} in DBpedia~\cite{DBpedia} and \textsf{Q513} in Wikidata~\cite{Wikidata}. Conventional approaches to this task exploit a wide range of discriminative features of entities, e.g., names, descriptive annotations, and relational structures~\cite{Alex,ObjectCoref,GenLink,SiGMa,PARIS}. The major challenge lies in the symbolic, linguistic and schematic heterogeneity between independently-created KGs.

\begin{figure}[t]
\centering
\includegraphics[width=0.75\columnwidth]{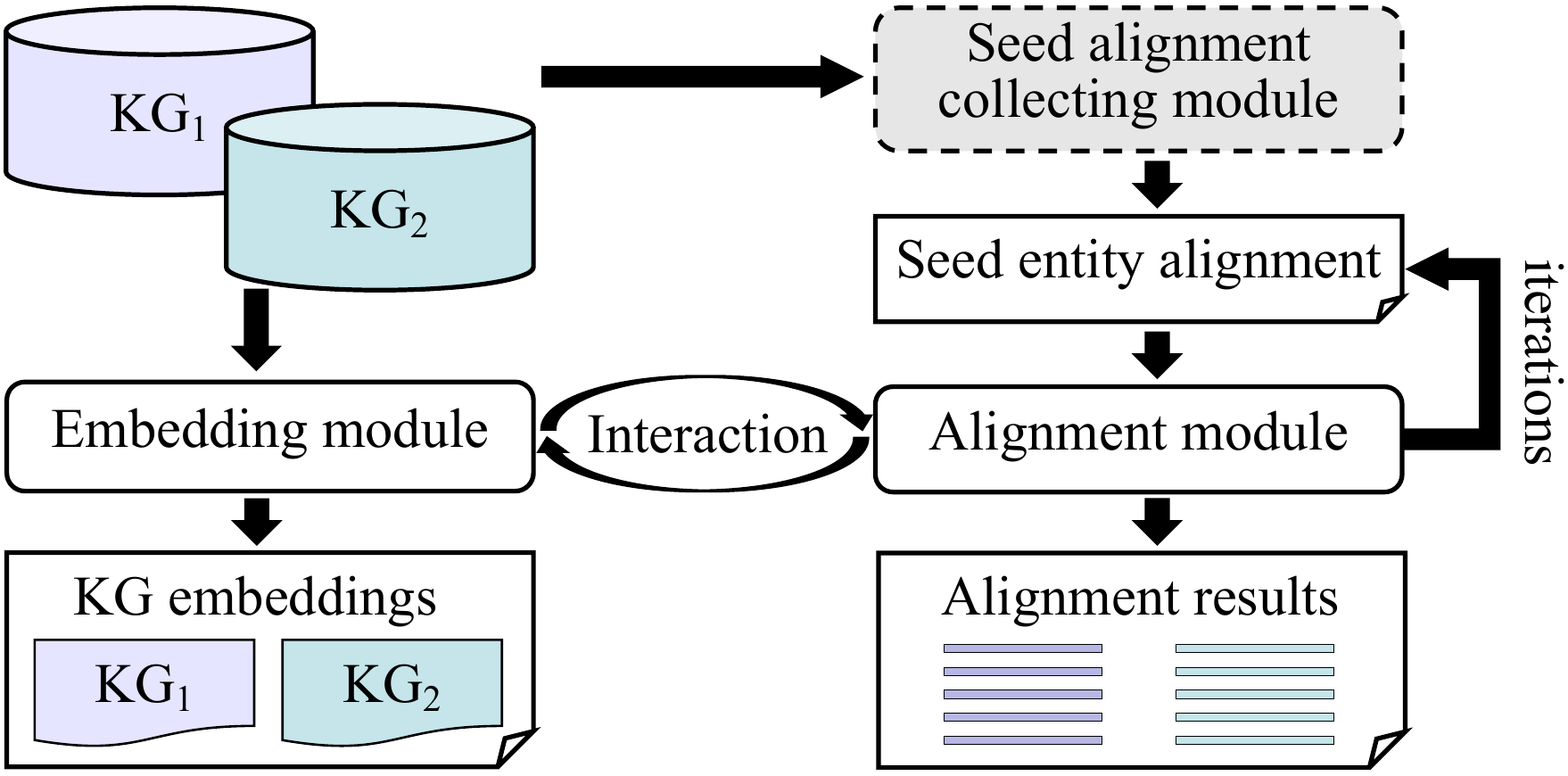}
\caption{Framework of embedding-based entity alignment}
\label{fig:framework}
\end{figure}

\emph{Embedding-based entity alignment} has emerged~\cite{MTransE} and seen much development in recent years~\cite{MuGNN,KDCoE,RSN,IMUSE,SEA,JAPE,BootEA,AttrE,GCNAlign,RDGCN,IPTransE}. This approach is based on KG embedding techniques, which embed the symbolic representations of a KG as low-dimensional vectors in a way such that the semantic relatedness of entities is captured by the geometrical structures of an embedding space \cite{TransE}. The premise is that such embeddings can potentially mitigate the aforementioned heterogeneity and simplify knowledge reasoning~\cite{KGESurvey}.~Figure~\ref{fig:framework} depicts a typical framework of embedding-based entity alignment. It takes as input two different KGs and collects \emph{seed alignment} between them using sources such as the \textsf{owl:sameAs} links \cite{MTransE}. Then, the two KGs and seed alignment are fed into the \emph{embedding} and \emph{alignment} modules, to capture the correspondence of entity embeddings. There are two typical combination paradigms for module interaction: (i) the embedding module encodes the two KGs in \emph{two} independent embedding spaces, meanwhile the alignment module uses seed alignment to learn a mapping between them \cite{KDCoE,MTransE,SEA,OTEA}; or (ii) the alignment module guides the embedding module to represent the two KGs into \emph{one} unified space by forcing the aligned entities in seed alignment to hold very similar embeddings \cite{MuGNN,NTAM,JAPE,BootEA,AttrE,GCNAlign,IPTransE}. Finally, entity similarities are measured by the learned embeddings. We can predict the counterpart of a source entity through the nearest neighbor search among target entity embeddings using a distance metric like the Euclidean distance. Besides, to overcome the shortage of seed entity alignment, several approaches \cite{KDCoE,BootEA,IPTransE} deploy semi-supervised learning to iteratively augment new alignment.

However, as an emerging research topic, there are still some issues with analyzing and evaluating embedding-based entity alignment. First, as far as we know, there is no prior work summarizing the status quo of this field yet. The \emph{latest development} of embedding-based entity alignment, as well as its \emph{advantages} and \emph{weaknesses} still remain to be explored. We even do not know how the embedding-based approaches compare to conventional entity alignment approaches. Second, there are also no widely-acknowledged benchmark datasets towards a realistic evaluation of embedding-based entity alignment. Arguably, a bit more popular datasets are DBP15K (used by \cite{MuGNN,KECG,MMEA,JAPE,BootEA,GCNAlign,RDGCN,HGCN,GMNN,HMAN,NAEA})) and WK3L (used by \cite{MTransE,NTAM,SEA,OTEA}). The different datasets for evaluation make it difficult to obtain a fair and comprehensive comparison of embedding-based entity alignment approaches. Moreover, current datasets contain much more high-degree entities (i.e., entities connected with many other entities, which are relatively easy for entity alignment) than real-world KGs do. As a result, many approaches may exhibit good performance on these biased datasets. Additionally, these datasets only focus on one aspect of heterogeneity, e.g., multilingualism, while overlook other aspects, e.g, different schemata and scales. This brings difficulties in understanding the generalization and robustness of embedding-based entity alignment. Third, we find that only a portion of the studies in this field come with source code, which makes it difficult to conduct further research on top of these approaches. Due to these issues, there is a pressing need to conduct a comprehensive and realistic re-evaluation of embedding-based entity alignment approaches with in-depth analysis.

In this paper, we carry out a systematic experimental study of embedding-based entity alignment with an open-source library. Our main contributions are listed as follows:
\begin{noindlist}
\item \emph{A comprehensive survey}. We survey 23 recent approaches for embedding-based entity alignment and categorize their core techniques and characteristics from different aspects. We also review the popular choices for each technical module, providing a brief overview of this field. (Sect.~\ref{sect:prelim})

\item \emph{Benchmark datasets}. To make a fair and realistic comparison, we construct a set of dedicated benchmark datasets with splits of five folds by sampling real-world KGs DBpedia \cite{DBpedia}, Wikidata \cite{Wikidata} and YAGO \cite{YAGO}, in consideration of various aspects of heterogeneity regarding entity degrees, multilingualism, schemata and scales. Particularly, we propose a new sampling algorithm, which can make the properties (e.g., degree distribution) of a sample approximate its source KG. (Sect.~\ref{sect:data})

\item \emph{Open-source library}. We develop an open-source library \emph{OpenEA}\footnote{\url{https://github.com/nju-websoft/OpenEA}\label{OpenEA}} using Python and TensorFlow. This library integrates 12 representative embedding-based entity alignment approaches belonging to a wide range of technologies. It uses a flexible architecture to make it easy to integrate a large amount of existing KG embedding models (8 representative ones have been implemented) for entity alignment. The library will be duly updated along with the coming of new approaches, to facilitate future research. (Sect.~\ref{sect:library})

\item \emph{Comprehensive comparison and analysis}. We provide a comprehensive comparison of 12 representative embedding-based entity alignment approaches in terms of both effectiveness and efficiency on our datasets. We train and tune each approach from scratch using our open-source library to ensure a fair evaluation. These results offer an overview of the performance of embedding-based entity alignment. To gain insights into the strengths and limitations of each approach, we conduct extensive analysis on their performance from different aspects. (Sect.~\ref{sect:exp})

\item \emph{Exploratory experiments}. We carry out three experiments beyond what has been available in literature. We give the first analysis on the geometric properties of entity embeddings to understand their underlying connections with the final performance. We notice that many KG embedding models have not been exploited for entity alignment and we explore 8 popular ones among them. We also compare embedding-based approaches with several conventional approaches, to explore their complementarity. (Sect.~\ref{sect:further_exp})

\item \emph{Future research directions}. Based on our survey and experimental findings, we provide a thorough outlook on several promising research directions for future work, including unsupervised entity alignment, long-tail entity alignment, large-scale entity alignment and entity alignment in non-Euclidean embedding spaces. (Sect.~\ref{sect:future})

\end{noindlist}

To the best of our knowledge, this work is the first systematic and comprehensive experimental study on embedding-based entity alignment between KGs. Our experiments reveal the true performance as well as the advantages and shortcomings of current approaches in the realistic entity alignment scenario. The shortcomings that we find, such as the incapacity of relation-based approaches in handling long-tail entities and the poor effectiveness of attribute-based approaches in resolving the heterogeneity of attribute values, call for the re-investigation of truly effective approaches for real-world entity alignment. We also believe that our in-depth analysis on the geometric properties of entity embeddings opens a new direction to investigate what enables the alignment-oriented embeddings and what supports the entity alignment performance behind the increasingly powerful approaches. Our benchmark datasets, library and experimental results are all publicly available through the GitHub repository\textsuperscript{\ref{OpenEA}} under the GPL license, to foster reproducible research. We think that the datasets and library will become a valuable and fundamental resource to future studies. As a growing number of knowledge-driven applications build their capacities on top of KGs and benefit from KG fusion, this work can lead to profound impacts to the KG and database communities.

\section{Preliminaries}
\label{sect:prelim}

We consider the entity alignment task between two KGs $\mathcal{KG}_1$ and $\mathcal{KG}_2$. Let $\mathcal{E}_1$ and $\mathcal{E}_2$ denote their entity sets, respectively. The goal is to find the 1-to-1 alignment of entities $\mathcal{S}_{\mathcal{KG}_1,\mathcal{KG}_2}=\{(e_1,e_2)\in\mathcal{E}_1\times\mathcal{E}_2 \,|\, e_1\sim e_2\}$, where $\sim$ denotes an equivalence relation \cite{SiGMa,PARIS}. In many cases, a small subset of the alignment $\mathcal{S}_{\mathcal{KG}_1, \mathcal{KG}_2}'\subset\mathcal{S}_{\mathcal{KG}_1, \mathcal{KG}_2}$, called \emph{seed alignment}, is known beforehand and used as training data. 

\subsection{Literature Review}
\label{subsect:review}

\subsubsection{Knowledge Graph Embedding}
\label{subsect:kge}

\noindent\textbf{Approaches.} Existing KG embedding models can be generally divided into three categories: (i) \emph{translational models}, e.g., TransE~\cite{TransE}, TransH~\cite{TransH}, TransR~\cite{TransR} and TransD~\cite{TransD}; (ii) \emph{semantic matching models}, e.g., DistMult~\cite{DistMult}, ComplEx~\cite{ComplEx}, HolE~\cite{HolE}, SimplE~\cite{SimplE}, RotatE \cite{RotatE} and TuckER~\cite{TuckER}; and (iii) \emph{deep models}, e.g., ProjE \cite{ProjE}, ConvE~\cite{ConvE}, R-GCN \cite{RGCN}, KBGAN~\cite{KBGAN} and DSKG~\cite{DSKG}. These models have been generally used for link prediction. We refer interested readers to the recent surveys \cite{KRLSurvey,KGESurvey}. A related area is network embedding \cite{RLGSurvey}, which learns vertex representations to capture their proximity. However, the edges in networks carry simplex semantics. This differentiates network embedding from KG embedding in both data models and learning techniques. 

\noindent\textbf{Datasets \& evaluation metrics.} FB15K and WN18 are two benchmark datasets for link prediction in KGs \cite{TransE}. Some studies notice that FB15K and WN18 suffer from the test leakage problem and build two new benchmark datasets FB15K-237 \cite{FB15K237} and WN18RR \cite{ConvE} correspondingly. Three metrics are widely used in evaluation: (i) proportion of correct links in the top-$m$ ranked results (called Hits@$m$, for example, $m=1$), (ii) mean rank (MR) of correct links, and (iii) mean reciprocal rank (MRR). Two efforts in evaluating link prediction models have been reported in \cite{AkramiSZHL20,2020KGE_Analysis}.

\subsubsection{Conventional Entity Alignment}
\label{subsect:cea}

\noindent\textbf{Approaches.} Conventional approaches address entity alignment mainly from two angles. One is based on \emph{equivalence reasoning} mandated by OWL semantics \cite{sameAs,LogMap}. The other is based on \emph{similarity computation}, which compares symbolic features of entities \cite{SiGMa,RiMOM,PARIS}. Recent studies also use statistical machine learning \cite{Alex,ObjectCoref,GenLink} and crowdsourcing \cite{Hike} to improve the accuracy. Also, in the database area, detecting duplicate entities, a.k.a. record linkage or entity resolution, has been extensively studied \cite{DRDSurvey,VLDBTutorial}. These approaches mainly rely on literal information of entities.

\noindent\textbf{Datasets \& evaluation metrics.} Since 2004, OAEI\footnote{\url{http://oaei.ontologymatching.org/}} (Ontology Alignment Evaluation Initiatives) has become the primary venue for work in ontology alignment. It also organizes an evaluation track for entity alignment in recent years. We have not observed any embedding-based systems participating in this track. The preferred evaluation metrics are precision, recall and F1-score. 

\subsubsection{Embedding-based Entity Alignment}
\label{subsect:eea}

\noindent\textbf{Approaches.} Many existing approaches \cite{MTransE,AKE,SEA,OTEA,JAPE,BootEA,AttrE,IPTransE} employ the translational models (e.g., TransE~\cite{TransE}) to learn entity embeddings for alignment based on relation triples. Some recent approaches \cite{MuGNN,KECG,GCNAlign,RDGCN,GMNN,HGCN,AVR-GCN,NAEA} employ graph convolutional networks (GCNs)~\cite{GCN,GAT}. Besides, some approaches incorporate attribute and value embeddings \cite{KDCoE,IMUSE,JAPE,AttrE,RDGCN,HGCN,HMAN,MultiKE}. We elaborate the techniques of these approaches in Sect.~\ref{subsect:stack}. Also, there are some approaches for (heterogeneous information) network alignment~\cite{REGAL,NTAM,MEgo2Vec} or cross-lingual knowledge projection \cite{LIN}, which may also be modified for entity alignment. It is also worth noting that two studies~\cite{DL4ER,Magellan} design the embedding-based approaches for entity resolution in databases. They represent the attribute values of entities based on word embeddings and compare entities using embedding distances. However, they assume that all entities follow the same schema or the attribute alignment must be 1-to-1 mapping. As different KGs are often created with different schemata, it is hard to fulfill these requirements. Thus, they cannot be applied to entity alignment of KGs. 

\begin{figure}
\centering
\includegraphics[width=\columnwidth]{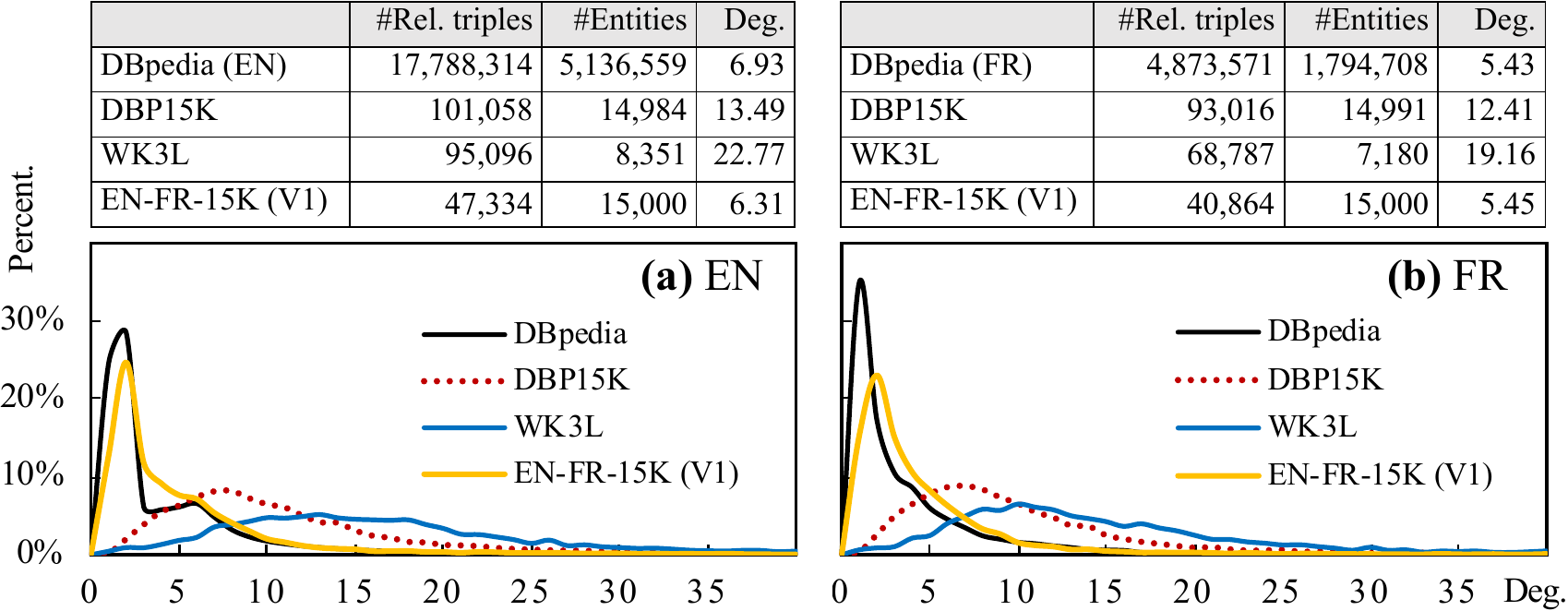}
\caption{Degree distributions and average degrees of two popular datasets DBP15K \protect\cite{JAPE} and WK3L \protect\cite{MTransE} used in previous approaches, along with our contributed dataset EN-FR-15K (V1). The x-axis denotes degrees and the y-axis denotes the percentage of entities w.r.t. degrees. These datasets are extracted from DBpedia \protect\cite{DBpedia}, but the degree distributions of DBP15K and WK3L are quite different from DBpedia and their average degrees are also larger. Our dataset retains a similar degree distribution to DBpedia.}
\label{fig:degree_current}
\end{figure}

\noindent\textbf{Datasets \& evaluation metrics.} To the best of our knowledge, there is no widely-acknowledged benchmark dataset for assessing embedding-based entity alignment approaches. Arguably, a bit more used datasets are DBP15K \cite{JAPE} and WK3L \cite{MTransE}. However, Figure~\ref{fig:degree_current} shows that their degree distributions and average degrees are significantly different from real-world KGs. More details about our datasets are reported in Sect.~\ref{sect:data}. Similar to link prediction, Hits@$m$, MR and MRR are mainly used as evaluation metrics, where Hits@1 should be emphasized, as it is equivalent to precision. 

\subsection{Categorization of Techniques}
\label{subsect:stack}

Table~\ref{tab:categ} categorizes 23 recent embedding-based entity alignment approaches by analyzing their embedding and alignment modules as well as the modes that they interact. For notations, we use capital calligraphic letters to denote sets and boldface letters for vectors and matrices.

\subsubsection{Embedding Module}
\label{subsect:embed_mod}

The embedding module seeks to encode a KG into a low-dimensional embedding space. Based on the types of triples used, we classify the KG embedding models in two types, i.e., \emph{relation embedding} and \emph{attribute embedding}. The former leverages relational learning techniques to capture KG structures, and the latter exploits attribute triples of entities.

\noindent\textbf{Relation embedding} is employed by all existing approaches. Below is three representative ways to realize it:

\emph{Triple-based embedding} captures the local semantics of relation triples. Many KG embedding models fall into this category, which defines an energy function to measure the plausibility of triples. For example, TransE~\cite{TransE} interprets a relation as the translation from its head entity embedding to its tail. The energy of a relation triple $(e_1, r_1, e_2)$ is
\begin{align} 
\label{eq:transe}
\phi(e_1, r_1, e_2) =\ \parallel\mathbf{e}_1 + \mathbf{r}_1 - \mathbf{e}_2\parallel,
\end{align}
where $\parallel\cdot\parallel$ denotes the $L_1$- or $L_2$-norm of vectors. TransE optimizes the \emph{marginal} ranking loss to separate positive triples from negatives by a pre-defined margin. Other choices of loss functions include the \emph{logistic} loss \cite{HolE,ComplEx} and the \emph{limit-based} loss \cite{BootEA,LimitLoss}. Negative triples can be generated using the \emph{uniform} negative sampling or \emph{truncated} sampling.

\emph{Path-based embedding} exploits the long-term dependency of relations spanning over relation paths. A relation path is a set of nose-to-tail linked relation triples, e.g., $(e_1, r_1, e_2), (e_2,$ $r_2, e_3)$. IPTransE \cite{IPTransE} models relation paths by inferring the equivalence between a direct relation and a multi-hop path. Assume that there is a direct relation $r_3$ from $e_1$ to $e_3$. IPTransE expects the embedding of $r_3$ to be similar to the path embedding, which is encoded as a combination of its constituent relation embeddings:
\begin{align} 
\label{eq:ptranse}
\mathbf{r}^* = \mathrm{comb}(\mathbf{r}_1, \mathbf{r}_2),
\end{align}
where $\mathrm{comb}(\cdot)$ is a sequence composition operation such as sum. $\parallel\mathbf{r}^*-\mathbf{r}_3\parallel$ is minimized to make them close to each other. However, IPTransE overlooks entities. Another work, RSN4EA~\cite{RSN}, modifies recurrent neural networks (RNNs) to model the sequence of entities and relations together.

\emph{Neighborhood-based embedding} uses the subgraph structure constituted by a large amount of relations between entities. GCNs \cite{Spectral,GraphCNN,GCN,RGCN} are well suited for modeling this structure, and have been used for embedding-based entity alignment recently \cite{MuGNN,KECG,GCNAlign,RDGCN,HGCN,GMNN,HMAN}. A GCN consists of multiple graph convolutional layers. Let $\mathbf{A}$ denote the adjacency matrix of a KG and $\mathbf{H}^{(0)}$ be a feature matrix where each row corresponds to an entity. The typical propagation rule from the $i^\text{th}$ layer to the ($i+1)^\text{th}$ layer \cite{GCN} is
\begin{align} 
\label{eq:gcn}
\mathbf{H}^{(i+1)} = \sigma(\hat{\mathbf{D}}^{-\frac{1}{2}} \hat{\mathbf{A}} \hat{\mathbf{D}}^{-\frac{1}{2}} \mathbf{H}^{(i)} \mathbf{W}),
\end{align}
with $\hat{\mathbf{A}}=\mathbf{A}+\mathbf{I}$ and $\mathbf{I}$ is an identity matrix. $\hat{\mathbf{D}}$ is the diagonal degree matrix of $\hat{\mathbf{A}}$. $\mathbf{W}$ is the learnable weight matrix. $\sigma(\cdot)$ is the activation function such as $\mathrm{tanh}(\cdot)$. 

\begin{table}
\centering
\caption{Categorization of popular embedding-based entity alignment approaches published before December 2019}
\label{tab:categ}
\begin{adjustbox}{width=\columnwidth}
\input{tab_categ}
\end{adjustbox}
\end{table}

\noindent\textbf{Attribute embedding} is used by several approaches \cite{KDCoE,IMUSE,JAPE,AttrE,GCNAlign,RDGCN,GMNN,HMAN,MultiKE} to enhance the similarity measure of entities. There are two ways for attribute embedding:

\emph{Attribute correlation embedding} considers the correlations among attributes. Attributes are regarded as correlated if they are frequently used together to describe an entity. For example, \textsf{longitude} is highly correlated with \textsf{latitude} as they often form a coordinate. JAPE~\cite{JAPE} exploits such correlations for entity alignment, based on the assumption that similar entities should have similar correlated attributes. For two attributes $a_1,a_2$, the probability that they are correlated is
\begin{align} 
\label{eq:attr}
\mathrm{Pr}(a_1,a_2) = \mathrm{sigmoid}(\mathbf{a}_1\cdot\mathbf{a}_2),
\end{align}
where attribute embeddings can be learned by maximizing the probability over all attribute pairs. Here, the attribute correlation embedding does not consider literal values.

\emph{Literal embedding} introduces literal values to attribute embedding. AttrE \cite{AttrE} proposes a character-level encoder that is capable of dealing with unseen values in training phases. Let $v=(c_1,c_2,...,$ $c_n)$ be a literal with $n$ characters, where $c_i$ ($1 \leq i \leq n$) is the $i^\text{th}$ character. AttrE~embeds~$v$~as
\begin{align} 
\label{eq:literal}
\mathbf{v} = \mathrm{comb}(\mathbf{c}_1, \mathbf{c}_2, \ldots, \mathbf{c}_n).
\end{align}
With this representation, literals are treated as entities and the relation embedding models like TransE can be used to learn from attribute triples. However, the character-based literal embedding may fail in cross-lingual settings.

\subsubsection{Alignment Module}
\label{subsect:align_mod}

The alignment module uses seed alignment as labeled training data to capture the correspondence of entity embeddings. Two keys are picking a distance metric and designing an alignment inference strategy.

\noindent\textbf{Distance metrics.} \emph{Cosine, Euclidean and Manhattan} distances are three widely-used metrics. In high-dimensional spaces, a few vectors (called hubs \cite{Hub}) may repeatedly occur as the $k$-nearest neighbors of others, the so-called \emph{hubness} problem \cite{WordTrans}. See Sect.~\ref{sect:geo_analysis} for more details.

\noindent\textbf{Alignment inference strategies.} \emph{Greedy search} is used by all current approaches. Given $\mathcal{KG}_1$ and $\mathcal{KG}_2$ to be aligned and a distance metric $\pi$, for each entity $e_1\in\mathcal{E}_1$, it finds the aligned entity $\tilde{e_2}$ by $\tilde{e_2}=\mathop{\arg\min}_{e_2\in\mathcal{E}_2} \pi(\mathbf{e}_1, \mathbf{e}_2)$. Differently, \emph{collective search} \cite{CollectiveSM,HolisticOM} aims to find a global optimal alignment that minimizes $\sum_{(e_1, e_2)\in\mathcal{S}_{\mathcal{KG}_1,\mathcal{KG}_2}} \pi(\mathbf{e}_1,\mathbf{e}_2)$. It can be modeled as the maximum weight matching problem in a bipartite graph and solved in $O(N^3)$ time using the Kuhn-Munkres algorithm ($N=|\mathcal{E}_1|+|\mathcal{E}_2|$), or reduced to linear time using the heuristic algorithm \cite{MWGM}. Another solution is the stable marriage algorithm \cite{SMP}. The alignment between $\mathcal{E}_1$ and $\mathcal{E}_2$ satisfies a stable marriage if there does not exist a pair of entities that both prefer each other than their current aligned ones. Its solution takes $O(N^2)$ time \cite{Gale_Shapley}.

\subsubsection{Interaction Mode}
\label{subsect:interact}

\noindent\textbf{Combination modes.} Four typical designs to reconcile KG embeddings for entity alignment are as follows: \emph{Embedding space transformation} embeds two KGs in different embedding spaces and learns a transformation matrix $\mathbf{M}$ between the two spaces using seed alignment, to achieve $\mathbf{M}\mathbf{e}_1 \approx\mathbf{e}_2$ for each $(e_1,e_2)\in\mathcal{S}'_{\mathcal{KG}_1,\mathcal{KG}_2}$. Another combination mode encodes two KGs into a unified embedding space. \emph{Embedding space calibration} minimizes $\parallel\mathbf{e}_1 - \mathbf{e}_2\parallel$ for each $(e_1,e_2)\in\mathcal{S}'_{\mathcal{KG}_1,\mathcal{KG}_2}$ to calibrate the embeddings of seed alignment. As two special cases, \emph{parameter sharing} directly configures $\mathbf{e}_1=\mathbf{e}_2$ and \emph{parameter swapping} swaps seed entities in their triples to generate extra triples as supervision. For instance, given $(e_1,e_2)\in\mathcal{S}'_{\mathcal{KG}_1,\mathcal{KG}_2}$ and a relation triple of $\mathcal{KG}_1$ $(e_1,r_1,e_1')$, parameter swapping produces a new triple $(e_2, r_1, e_1')$ and feeds it in KG embedding models as a real triple. Both parameter sharing and swapping methods do not introduce new loss functions, but the latter produces more triples.

\noindent\textbf{Learning strategies.} Based on how to process labeled and unlabeled data, learning strategies can be divided below:

\emph{Supervised learning} leverages the seed alignment as labeled training data. For embedding space transformation, seed alignment is used to learn the transformation matrix. For space calibration, it is used to let aligned entities have similar embeddings. But, the acquisition of seed alignment is costly and error-prone, especially for cross-lingual KGs.

\emph{Semi-supervised learning} uses unlabeled data in training, e.g., self-training \cite{BootEA,IPTransE} and co-training \cite{KDCoE}. The former iteratively proposes new alignment to augment seed alignment. The latter combines two models learned from disjoint entity features and alternately enhances the alignment learning of each other. Although OTEA~\cite{OTEA} and KECG~\cite{KECG} claim that they are semi-supervised approaches, their learning strategies do not augment seed alignment. We do not treat them as standard semi-supervised learning in this paper.

\emph{Unsupervised learning} needs no training data. We have not observed any embedding-based entity alignment approaches using unsupervised learning. Although IMUSE~\cite{IMUSE} claims that it is an unsupervised approach, it actually uses a preprocessing method to collect seed alignment with high string similarity. Its embedding module still needs seed alignment. 

\section{Dataset Generation}
\label{sect:data}

As aforementioned, current widely-used datasets are quite different from real-world KGs. Also, it is hard for embedding-based approaches to run on full KGs due to the large and unpartitioned candidate space. Hence, we sample real-world KGs and provide two data scales (15K and 100K).

\begin{algorithm}[!t]
	\caption{Iterative degree-based sampling (IDS)}
	\label{algo:IDS}
	{\small
		\KwIn{$\mathcal{KG}_1, \mathcal{KG}_2$, reference alignment $\mathcal{S}_{ref}$, entity size $N$, hyper-parameters $\mu,\epsilon$}
		\SetKwRepeat{Do}{do}{while}
		\tcp{only retain entities in reference alignment}
		Filter $\mathcal{KG}_1, \mathcal{KG}_2$ by $\mathcal{S}_{ref}$\;
		Get degree distributions $Q_1,Q_2$ for $\mathcal{KG}_1,\mathcal{KG}_2$, resp.\;
		\Do( \tcp*[h]{if fails, run it again}){$JS(Q_1,P_1)>\epsilon \parallel JS(Q_2,P_2)>\epsilon$}{
			Initialize datasets $\mathcal{DS}_1,\mathcal{DS}_2$ from $\mathcal{KG}_1,\mathcal{KG}_2$, resp.\;
			\While{$|\mathcal{DS}_1|>N\ \&\&\ |\mathcal{DS}_2|>N$}{
				\For{$\mathcal{DS}_j\ (j=1,2)$}{
					Get $dsize_j(x,\mu)$ for each degree $x$\;
					Get entity deletion probability by PageRank\;
					Delete $dsize_j(x,\mu)$ entities w.r.t.~probabilities\;}
				Filter $\mathcal{DS}_1,\mathcal{DS}_2$ by $\mathcal{S}_{ref}$; update $\mathcal{S}_{ref}$ accordingly\;}
			Get degree distributions $P_1,P_2$ for $\mathcal{DS}_1,\mathcal{DS}_2$, resp.\;}
		\Return $\mathcal{DS}_1,\mathcal{DS}_2,\mathcal{S}_{ref}$\;}
\end{algorithm}

\begin{table*}[!t]
\centering
\caption{Dataset statistics}
\label{tab:stats}
\begin{adjustbox}{width=\textwidth}
\input{tab_stats}
\end{adjustbox}
\vspace{-10pt}
\end{table*}

\subsection{Iterative Degree-based Sampling}
\label{subsect:sampling}

We consider five factors in building our datasets: source KGs, reference alignment, dataset sizes, languages and density, where the last is more challenging for building datasets. Specifically, we want to generate a certain-sized dataset from a source KG such that the difference of their entity degree distributions does not exceed an expectation. The difficulty lies in that the removal of an entity from the source KG also changes the connectivity of its neighboring entities. 

We propose an iterative degree-based sampling (IDS) algorithm, which simultaneously deletes entities in two source KGs with reference alignment until achieving the desired size, meanwhile keeping a similar degree distribution of each sampled dataset as the source KG. Algorithm~\ref{algo:IDS} describes the sampling procedure. During iterations, the proportion of entities having degree $x$ in the current dataset, denoted by $P(x)$, cannot always equal the original proportion $Q(x)$. We adjust the entity size to be deleted by $dsize(x,\mu)=\mu\big(1+P(x)-Q(x)\big)$, where $\mu$ is the base step size (see Line 7). Moreover, we prefer not to delete entities having a big influence on the overall degree distribution, such as the ones of high degree. To achieve this, we leverage the PageRank value for measuring the probability of an entity to be deleted (Line 8).

We use the Jensen-Shannon (JS) divergence \cite{JS} to assess the difference of two degree distributions (Line 12). Given two degree distributions $Q,P$, their JS-divergence is:
\begin{align} 
\label{eq:JS}
JS(Q,P)=\frac{1}{2}\sum_{x=1}^{n} \Big( Q(x)\log\frac{Q(x)}{M(x)} + P(x)\log\frac{P(x)}{M(x)} \Big),
\end{align}
where $Q(x)$ and $P(x)$ denote the proportions of entities with degree $x$ ($x=1\ldots n$) in $Q,P$, respectively, and $M=\frac{Q+P}{2}$. A small JS divergence $\epsilon$ between $Q$ and $P$ reveals that they have similar degree distributions. We set expectation $\epsilon\leq5\%$. The most costly part of IDS is to calculate PageRank weights during the iterations of deleting entities. It can be scaled to very large KGs by using approximation algorithms~\cite{NewBib_FastPageRank}.

\subsection{Dataset Overview}
\label{subsect:dataset}

We choose three well-known KGs as our sources: DBpedia (2016-10) \cite{DBpedia}, Wikidata (20160801) \cite{Wikidata} and YAGO~3~\cite{YAGO}. Also, we consider two cross-lingual versions of DBpedia: English--French and English--German. We follow the conventions in \cite{MTransE,JAPE,BootEA,GCNAlign,IPTransE} to generate datasets of two sizes with 15K and 100K entities, using the IDS algorithm. Specifically, we make use of DBpedia's inter-language links and \textsf{owl:sameAs} among the three KGs to retrieve reference entity alignment. To balance the efficiency and deletion safety, we set $\mu=100$ for 15K and $\mu=500$ for 100K.

\begin{figure}
\centering
\includegraphics[width=\columnwidth]{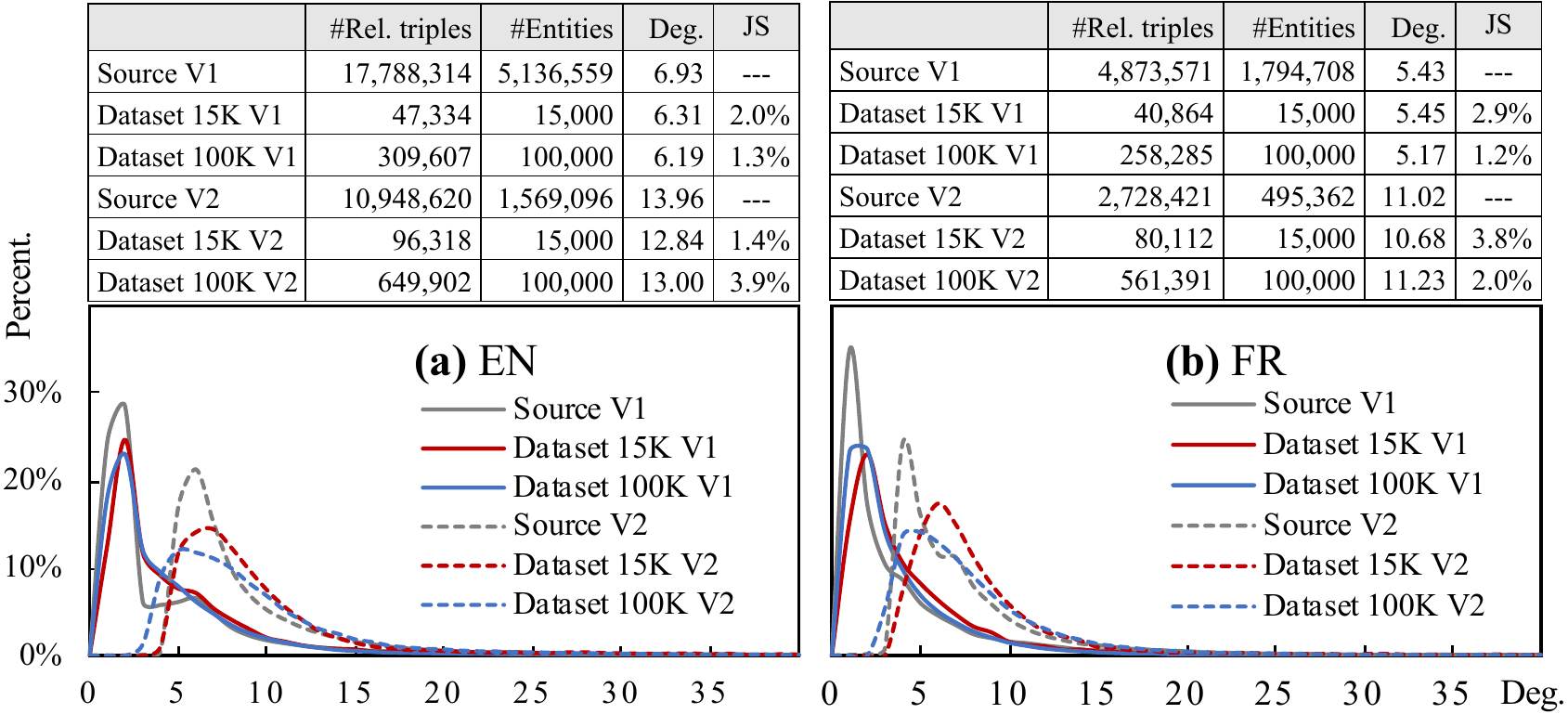}
\caption{{Degree distributions and average degrees of our sampled datasets EN-FR-15K (V1, V2) and EN-FR-100K (V1, V2), compared with DBpedia (the source KG)}}
\label{fig:degree_CDF}
\end{figure}

The statistics of the datasets are listed in Table \ref{tab:stats}. We generate two versions of datasets for each pair of source KGs.  V1 is gained by directly using the IDS algorithm. For V2, we first randomly delete entities with low degrees ($d\leq 5$) in the source KG to make the average degree doubled, and then execute IDS to fit the new KG. As a result, V2 is twice denser than V1 and more similar to existing datasets \cite{MTransE,JAPE}. Figure~\ref{fig:degree_CDF} shows the degree distributions and average degrees of EN-FR-15K (V1, V2) and EN-FR-100K (V1, V2). Our 15K and 100K datasets are much closer to the source KGs.

For each dataset, we also extract the attribute triples of entities to fulfill the input requirement of some approaches \cite{KDCoE,IMUSE,JAPE,AttrE,GCNAlign,RDGCN,GMNN,MultiKE}. Considering that DBpedia, Wikidata and YAGO collect data from very similar sources (mainly, Wikipedia), the aligned entities usually have identical labels. They would become ``tricky'' features for entity alignment and influence the evaluation of real performance. According to the suggestion in \cite{PBA}, we delete entity labels. 

By convention, we split a dataset into training, validation and test sets. The details are given in  Sect.~\ref{subsect:setting}. 

\subsection{Dataset Evaluation}
\label{subsect:dataset_eval}
We assess IDS and the quality of our datasets. Note that, generating an entity alignment dataset is a non-trivial work, as a qualified dataset needs to hold several characteristics, such as good connectivity (due to many approaches rely on graph structures), similar degree distributions to original KGs (for a realistic entity alignment scenario), and enough alignment (for training/validation/test). As far as we know, there still lacks a sampling method dedicated to this problem. For evaluation, we design two baseline methods on the basis of existing graph sampling algorithms \cite{Sampling}:
\begin{noindlist}
\item \emph{Random alignment sampling (RAS)} first randomly selects a fixed size (e.g., 15K) of entity alignment between two KGs, and then extracts the relation triples whose head and tail entities are both in the sampled entities.
\item \emph{PageRank-based sampling (PRS)} first samples entities from one KG based on the PageRank scores (entities not involved in any alignment are discarded), and then extracts these entities' counterparts from the other KG.
\end{noindlist}

\begin{table}
\centering
\caption{{Comparison of the EN-FR-15K (V1) datasets generated by RAS, PRS and IDS}}
\label{tab:data_evaluation}
\begin{adjustbox}{width=0.99\columnwidth}
\begin{tabular}{|c|c|c|c|c|c|c|}
	\hline \rowcolor{gray!10} {Datasets} & {KGs} & {\#Alignment}& {Deg.} & {JS} & {Isolates} & {Cluster coef.}\\ 
	\hline \multirow{2}{*}{{DBpedia}} 
	& {EN} & \multirow{2}{*}{{525,807}} & {6.39} & {--} & {0} & {0.342}\\ 
	& {FR} &  & {5.43} & {--} & {0}  & {0.080}\\
	\hline \multirow{2}{*}{{RAS}} 
	& {EN} & \multirow{2}{*}{{\,~15,000}} & {0.27} & {14.5\%} & {85.5\%} & {0.002}\\ 
	& {FR} &  & {0.17} & {12.1\%} & {90.1\%} & {0.001}\\
	\hline \multirow{2}{*}{{PRS}} 
	& {EN} & \multirow{2}{*}{{\,~15,000}}& {1.20} & {\,~7.3\%} & {68.9\%} & {0.025}\\ 
	& {FR} &  & {0.63} & {\,~9.3\%} & {69.4\%} & {0.015}\\
		\hline \multirow{2}{*}{{IDS}} 
	& {EN} & \multirow{2}{*}{{\,~15,000}}& {6.31} & {\,~2.0\%} & {0} & {0.233}\\ 
	& {FR} &  & {5.45} & {\,~2.9\%} & {0} & {0.190}\\
	\hline
\end{tabular}
\end{adjustbox}
\end{table}

Table~\ref{tab:data_evaluation} lists the properties of EN-FR-15K (V1) datasets generated by RAS, PRS and our IDS, compared to the source KGs (relation triples). In addition to the average degree and JS-divergence, we further consider two metrics: percentage of isolated entities~\cite{NewBib_Obj_link_struct} and clustering coefficient~\cite{Sampling}. The dataset of RAS is much sparser than the source, because the random sampling cannot retain the connectivity and degree distribution \cite{NewBib_Stumpf4221}. It has a low clustering coefficient and contains many isolated entities that are typically hard for embedding modules to handle. PRS more focuses on high-degree entities and gets better properties than RAS. However, the dataset is still far away from satisfactory due to the low average degree, high JS value and high percentage of isolated entities. This is because its entity selection procedure only applies to one KG rather than two KGs together. Differently, IDS considers the degree distributions of two KGs together. It tends to sample two aligned entities with similar degrees. Thus, the two KGs of our dataset have similar clustering coefficients. As the sampled dataset is much smaller that the source, it is hard to keep all these properties well. IDS shows good comprehensive performance.

\section{Open-source Library}
\label{sect:library}
\begin{figure}
	\centering
	\setlength{\belowcaptionskip}{1pt}
	\includegraphics[width=\columnwidth]{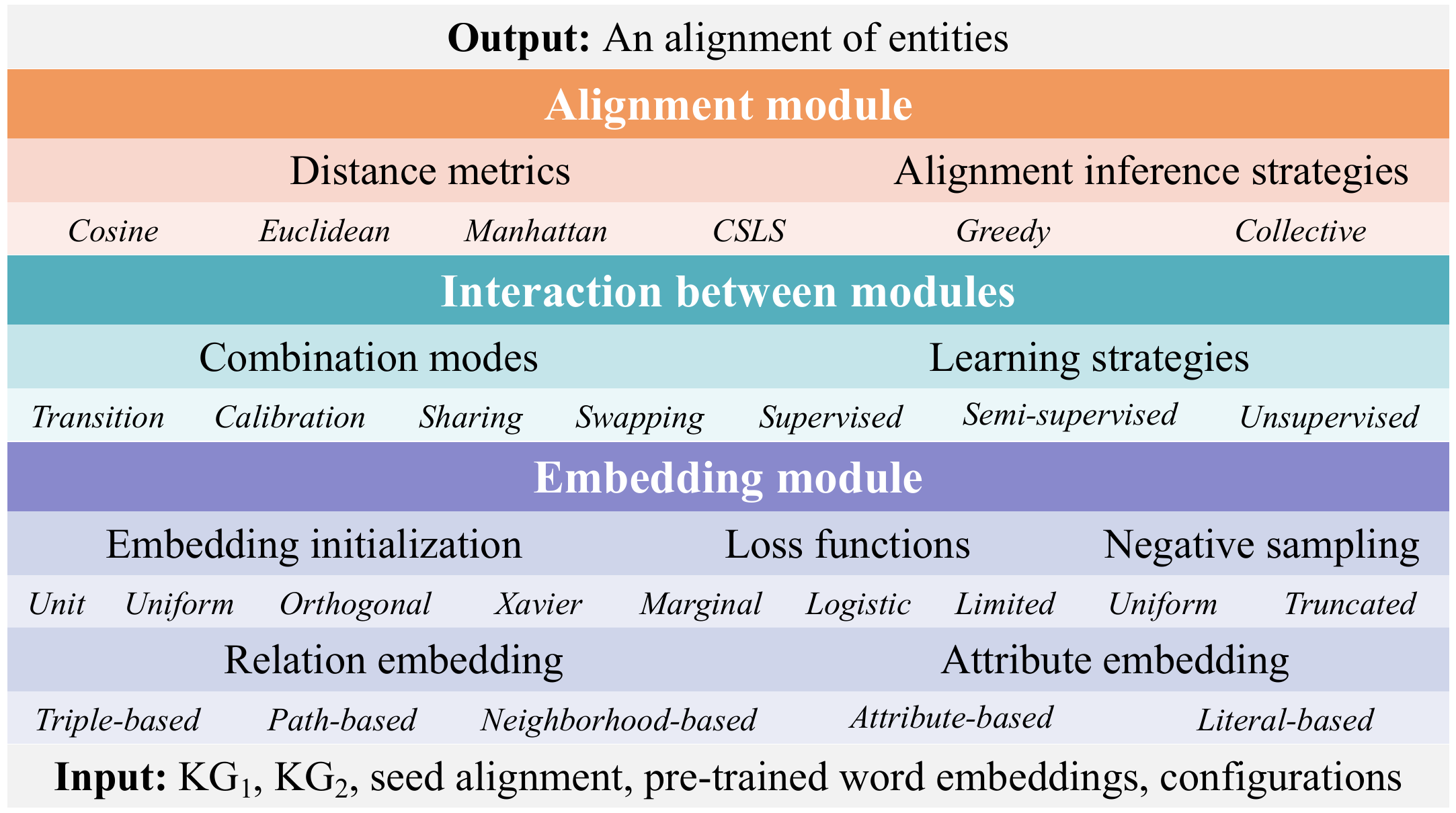}
	\caption{Software architecture of OpenEA}
	\label{fig:stack}
\end{figure}

We use Python and TensorFlow to develop an open-source library, namely \textbf{OpenEA}, for embedding-based entity alignment. The software architecture is illustrated in Figure~\ref{fig:stack}. Our design goals and features include three aspects:

\noindent\textbf{Loose coupling.} 
The implementation of embedding and alignment modules is independent to each other. OpenEA provides a framework template with pre-defined input and output data structures to make these modules as an integral pipeline. Users can freely call and combine different techniques in these modules to develop new approaches.

\noindent\textbf{Functionality and extensibility.} 
OpenEA implements a set of necessary functions as its underlying components, including initialization functions, loss functions and negative sampling methods in the embedding module; combination and learning strategies in the interaction mode; as well as distance metrics and alignment inference strategies in the alignment module. On top of those, OpenEA also provides a set of flexible and high-level functions with configuration options to call these components. In this way, new functions can be easily integrated by adding new configuration options.

\noindent\textbf{Off-the-shelf approaches.} 
To facilitate the usage of OpenEA and support our experimental study, we try our best to integrate or rebuild 12 representative embedding-based entity alignment approaches belonging to a wide range of technologies, including MTransE, IPTransE, JAPE, KDCoE, BootEA, GCNAlign, AttrE, IMUSE, SEA, RSN4EA, MultiKE and RDGCN. MTransE, JAPE, KDCoE, BootEA, GCNAlign, AttrE, RSN4EA, MultiKE and RDGCN are implemented by integrating their source code, while IPTransE, IMUSE and SEA are rebuilt by ourselves. Moreover, we integrate several relation embedding models that have not been explored for entity alignment yet, including three translational models TransH~\cite{TransH}, TransR \cite{TransR} and TransD \cite{TransD}; three semantic matching models HolE \cite{HolE}, SimplE \cite{SimplE} and RotatE~\cite{RotatE}; as well as two deep models ProjE~\cite{ProjE} and ConvE \cite{ConvE}. We also integrate two attribute embedding models AC2Vec \cite{JAPE} and Label2Vec \cite{MultiKE}, based on pre-trained multilingual word embeddings~\cite{fastText}. TransH, TransR, TransD and HolE are developed by referring to the open-source toolkit OpenKE \cite{OpenKE}; the remaining is implemented based on their source code.

\section{Experiments and Results}
\label{sect:exp}

In this section, we report a comprehensive evaluation using our benchmark datasets and open-source library.

\subsection{Experiment Settings}
\label{subsect:setting}

\noindent\textbf{Environment.} We carry out the experiments on the workstation with an Intel Xeon E3 3.3GHz CPU, 128GB memory, a NVIDIA GeForce GTX 1080Ti GPU and Ubuntu 16.04.

\noindent\textbf{Cross-validation.} We conduct the experiments with 5-fold cross-validation to ensure unbiased evaluation. Specifically, we divide the reference entity alignment into five disjoint folds, each of which accounts for 20\% of the total. For each running, we pick one fold (20\%) as training data and leave the remaining for validation (10\%) and testing (70\%). As found in \cite{MTransE}, the inter-language links in the multilingual Wikipedia cover about 15\% of entity alignment. Thus, using $20\%$ as training data can both satisfy the need for 5-fold cross-validation and conform to the real world.

\begin{table}
	\centering
	\caption{Common hyper-parameters for all the approaches}
	\label{tab:param}
	\begin{adjustbox}{width=\columnwidth}
		\begin{tabular}{|l|>{\centering}p{3.3cm}|>{\centering}p{3.3cm}|}
			\hline \rowcolor{gray!10} & 15K & 100K \tabularnewline 	
			\hline Batch size for rel. trip. & 5,000 & 20,000 \tabularnewline 
			\hline \multirow{2}{*}{Termination condition} & \multicolumn{2}{l|}{Early stop when the Hits@1 score begins to drop} \tabularnewline
			& \multicolumn{2}{l|}{on the validation sets, checked every 10 epochs.} \tabularnewline
			\hline Max. epochs & \multicolumn{2}{c|}{2000} \tabularnewline 
			\hline
		\end{tabular}
	\end{adjustbox}
\end{table}

\noindent\textbf{Comparative approaches and settings.} We evaluate all the embedding-based entity alignment approaches implemented in OpenEA. To make a fair comparison, we use our best efforts to unify the experiment settings. Table~\ref{tab:param} shows the common hyper-parameters used for all the approaches. As indicated in \cite{StrikeBack}, the batch size has an influence on the performance and running time. So, we use a fixed batch size for relation triples to avoid its interference. For other settings specific to each approach, we follow the reported details in literature as carefully as we can, {e.g., the margin for the ranking loss in IPTransE and AttrE is $1.5$; the number of GCN layers in GCNAlign and RDGCN is $2$.}  For several key hyper-parameters and the unreported ones, we try our best to tune them. For example, we constrain the $L_2$-norm of entity embeddings to 1 for many approaches, e.g., IMUSE, because we find that such normalization yields better results. For cross-lingual datasets, we use pre-trained cross-lingual word embeddings \cite{fastText} to initialize literal embeddings for the approaches using attribute values. The hyper-parameter settings of each approach on our datasets are available online. Notice that there are emerging approaches (e.g., AliNet \cite{AliNet}) that are contemporaneous to this paper. We will accordingly include those approaches into future release of OpenEA.

\noindent\textbf{Evaluation metrics.} In our experiments, the default alignment direction is from left to right. Take D-W for example. We treat DBpedia as the source and align it with the target KG Wikidata. Following the conventions, we use Hits@$m$ ($m=1,5$), MR and MRR as the evaluation metrics.

\noindent\textbf{Availability.} We release the datasets and OpenEA library online. The experimental results on five folds of each dataset using all the metrics are provided in the CSV format. All will be duly updated along with the coming of new approaches.

\subsection{Main Results and Analysis}
\label{subsect:main}

Table~\ref{tab:main_results} depicts the Hits@1, Hits@5 and MRR results of the 12 implemented approaches on our datasets. In summary, RDGCN, BootEA and MultiKE achieve the top-3 results. For a comprehensive and thorough understanding, we analyze the results from {five} angles:

\begin{table*}[!t]
\centering
\caption{Cross-validation results of current representative approaches on the 15K and 100K datasets}
\label{tab:main_results}
\begin{adjustbox}{width=\textwidth}
\input{tab_main_results}
\end{adjustbox}
\vspace{-10pt}
\end{table*}

\begin{figure}[!t]
	\centering
	\includegraphics[width=0.95\linewidth]{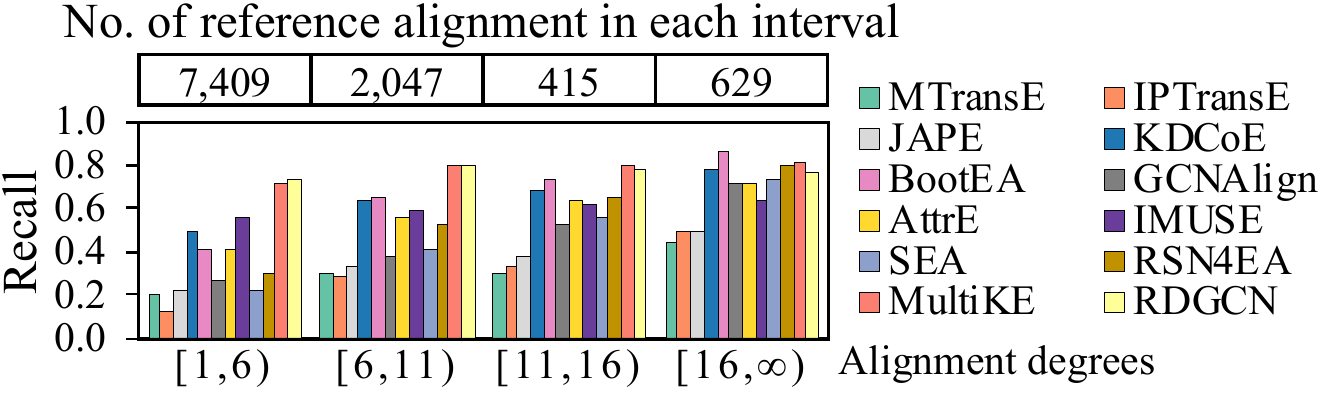}
	\centering\caption{Recall w.r.t.~alignment deg. on EN-FR-15K\,(V1)}
	\label{fig:degree_recall}
\end{figure}

\noindent\textbf{Sparse datasets (V1) vs.~dense datasets (V2).} From Table~\ref{tab:main_results}, we find that most relation-based approaches perform better on the dense datasets than on the sparse ones, e.g., IPTransE, BootEA, SEA and RSN4EA. This is in accord with our intuition that the entities in the dense datasets are generally involved in more relation triples, which enable these approaches to capture more semantic information. For the approaches considering attribute triples, KDCoE, GCNAlign, AttrE, IMUSE and RDGCN also perform better on the dense datasets, indicating that the relation embeddings still make contributions. Differently, MultiKE relies on multiple ``views" of features, which make it relatively insensitive to the relation changes. Interestingly, we also see that the performance of two relation-based approaches, MTransE and JAPE, drops on some dense datasets. We believe that this is because they are based on TransE, which has deficiency in handling multi-mapping relations in the dense datasets. {For example, $39.0\%$ of entities in EN-FR-100K (V1) have multi-mapping relations while the proportion in EN-FR-100K (V2) reaches up to $71.2\%$.} The complex structures make MTransE and JAPE prone to learn very similar embeddings for different entities involving the same multi-mapping relation \cite{TransR,TransH}.

For further analysis, we divide the test alignment of each dataset into multiple groups in terms of alignment degrees. The degree of an alignment is defined as the sum of relation triples for the two involved entities. Figure \ref{fig:degree_recall} illustrates the recall results on EN-FR-15K (V1). Obviously, most entities have relatively few relation triples, and we call them \emph{long-tail} entities. {\ul{\emph{We find that all the relation-based approaches run better in aligning entities with rich relation triples while their results decline on long-tail entities}}, as long-tail entities have little information useful for learning, which limits the expressiveness of their embeddings.} This lopsided performance confirms the results on the sparse and dense datasets from another angle. By using additional literals, the lopsided performance of KDCoE, AttrE, IMUSE, MultiKE and RDGCN alleviates. However, JAPE and GCNAlign that use attribute correlations still show the lopsided performance for entities with different degrees. The experiments on other datasets also agree on the above observations. Currently, we have not seen a method that handles long-tail entities well.

\noindent\textbf{15K datasets vs.~100K datasets.} We observe that all the approaches perform better on the 15K datasets than on the 100K datasets, except D-Y, {because the 100K datasets have more complex structures, causing more difficulties for embedding-based approaches to capture entity proximity. For example, $34.9\%$ of entities in EN-FR-15K (V1) are involved in multi-mapping relations while the proportion in EN-FR-100K (V1) reaches $39.0\%$. As we have discussed, multi-mapping relations challenge many embedding approaches. Moreover, the 100K datasets have a larger candidate alignment space than the 15K datasets. It is harder to rank the target entity at the top from a larger candidate space with much more negative cases.} Differently, D-Y-15K and D-Y-100K have a very similar number of relations in YAGO, which makes the results different from those on other datasets.

\begin{figure}[t]
	\centering
	\includegraphics[width=\linewidth]{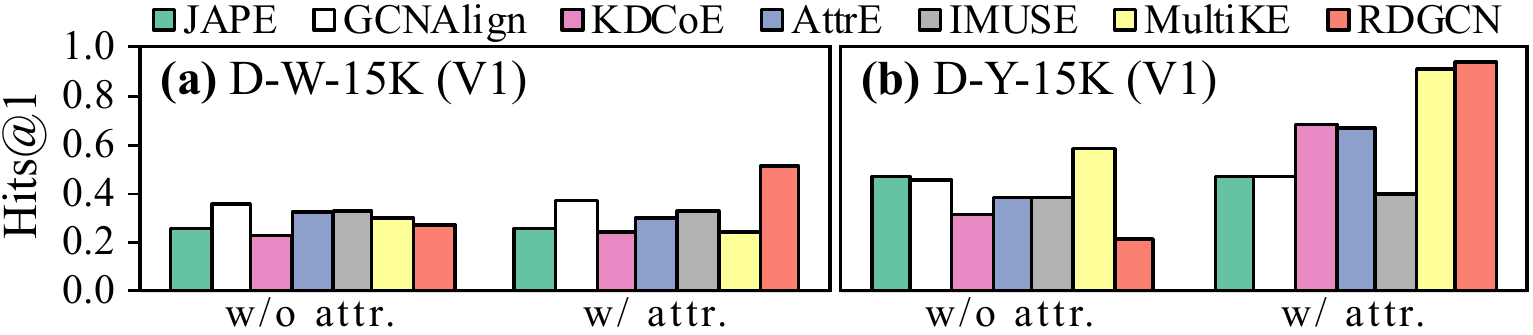}
	\centering\caption{Hits@1 results of JAPE, GCNAlign, KDCoE, AttrE, IMUSE, MultiKE, RDGCN and their degraded variants without attribute embedding}
	\label{fig:wo_attr}
\end{figure} 

\begin{figure}
	\centering
	\includegraphics[width=\linewidth]{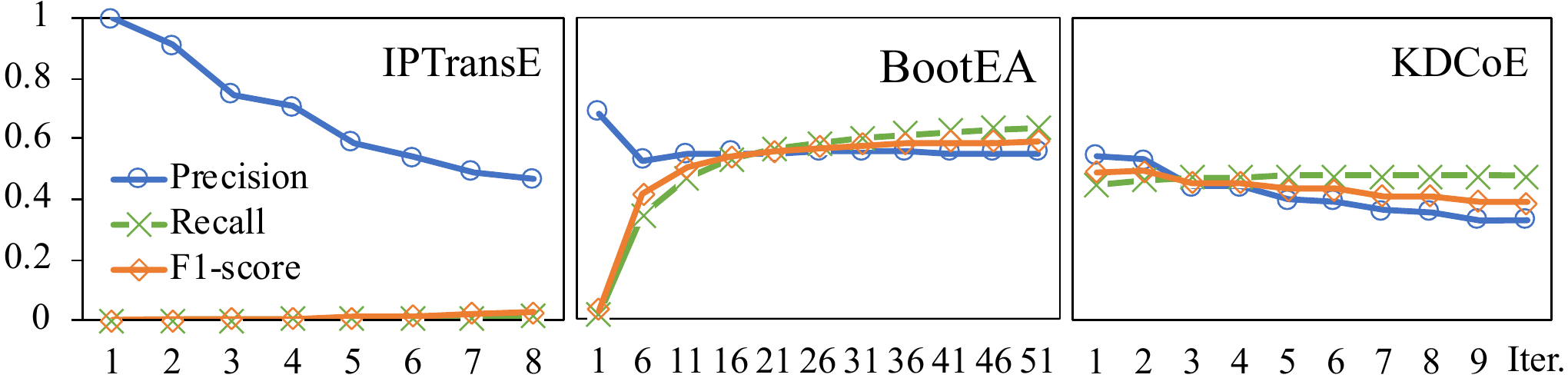}
	\centering\caption{Precision, recall and F1-score of augmented alignment during iterations on EN-FR-100K (V1)}
	\label{fig:semi_iter}
\end{figure}

\noindent\textbf{Relations vs.~attributes.} For the purely relation-based approaches, there is no clear advantage of one relation embedding technique beyond another. For example, although MTransE and BootEA both use TransE, their performance is at two extremes. We believe that the negative sampling in BootEA makes great contribution, {and training embeddings only with positive samples is prone to overfitting.} The work in \cite{KBGAN} also shows that negative sampling can largely affect the expressiveness of KG embeddings. {We apply the negative sampling along with the marginal ranking loss to MTransE and find that its Hits@1 on EN-FR-15K (V1) rises to $0.271$, which further demonstrates the effectiveness of negative sampling.} Besides, the bootstrapping strategy of BootEA also contributes a lot, which is discussed shortly. For another example, IPTransE and RSN4EA both extend triple-based embedding by linking relation triples into long relation paths, but their results are also significantly different. This is because the recurrent skipping network of RSN4EA is more powerful than the shallow composition of IPTransE. 

For the approaches using attributes, we compare them to their variants without attribute embedding. Figure~\ref{fig:wo_attr} shows the Hits@1 results on D-W-15K (V1) and D-Y-15K (V1). Other datasets show similar results. On D-Y, we do not observe notable improvement from JAPE and GCNAlign by using attribute correlations to cluster entities. This technique would fail to capture the attribute correlations across different KGs without pre-aligned attributes. Moreover, even if the attribute correlations are discovered, this signal is too coarse-grained to determine whether two entities with correlated attributes are aligned. Differently, literal embedding brings significant improvement to most approaches except IMUSE, indicating that literals are a stronger signal for entity alignment than attribute correlations. IMUSE has a preprocessing step using literals to find new entity alignment to augment training data. However, the errors in new alignment also harm performance. Most approaches fail to be improved by attribute embedding on D-W. The symbolic heterogeneity of attributes in Wikidata (e.g., the local names of attributes are numeric IDs) notably challenges some approaches as they cannot automatically find high-quality attribute alignment for literal comparison. {\ul{\emph{Overall, attribute heterogeneity has a strong effect on capturing attribute correlations, and literal embedding facilitates entity alignment.}}}

\begin{figure}[t]
	\centering
	\includegraphics[width=0.99\linewidth]{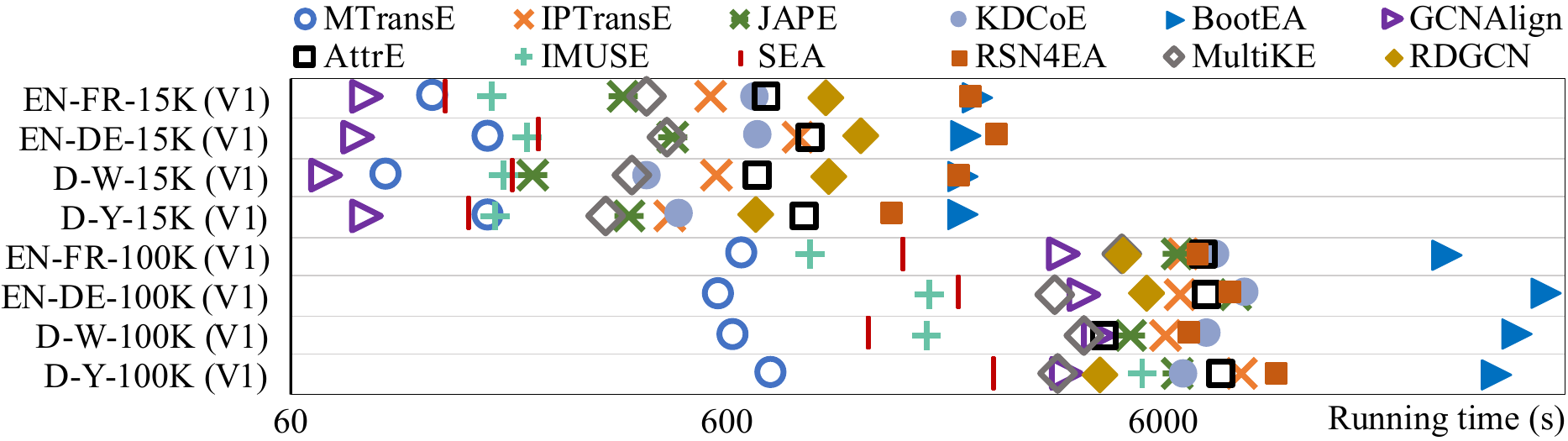}
	\centering\caption{{Running time (in log scale) on the V1 datasets}}
	\label{fig:time}
\end{figure} 

\noindent\textbf{Semi-supervised learning strategies.} 
We further investigate the strengths and limitations of these semi-supervised learning strategies by analyzing the quality of the augmented seed alignment. Figure~\ref{fig:semi_iter} depicts the precision, recall and F1-score of IPTransE, BootEA and KDCoE during the semi-supervised training on EN-FR-100K (V1), and other datasets show similar results. IPTransE fails to achieve good performance, because it involves many errors as the self-training continues but does not design a mechanism to eliminate these errors. KDCoE propagates new alignment by co-training two orthogonal types of features, i.e., relation triples and textual descriptions. {However, many entities lack textual descriptions, preventing KDCoE from 
finding alignment seeds to augment training data. Thus, its strategy does not bring notable improvement.} BootEA employs a heuristic editing method to remove wrong alignment. After undergoing a period of fluctuations, the precision stays stable while the recall continues growing during self-training, which brings a clear performance boost. {We also conduct an ablation study on BootEA, and find that its self-training strategy can bring an improvement of more than $0.086$ Hits@1 on the V1 datasets, demonstrating its effectiveness. {\ul{\emph{So, the quantity and quality of the augmented entity alignment have great impact on the semi-supervised approaches.}}} A larger augmented alignment of higher precision leads to better performance.}

\noindent\textbf{{Running time comparison.}}
{In Figure~\ref{fig:time}, we show a brief comparison on the average running time of five repetitions on the V1 datasets. The time used by different approaches varies greatly. In general, an approach takes more time to run on a 100K dataset than on a 15K dataset. BootEA is much slower than other approaches. For example, its running time on EN-FR-15K (V1) and EN-FR-100K (V1) are 2,260 and 26,939 seconds, respectively, where the truncated negative sampling and bootstrapping procedure cost more than $23.5\%$ and $13.3\%$ of the time, respectively. RSN4EA also uses much time, especially on the 15K (V1) datasets, since it is trained with multi-hop paths, which are far more than the relation triples (i.e., one-hop paths). For instance, the number of two-hop paths in EN-FR-15K (V1) is 500,260, five times more than that of relation triples (88,198). As for KDCoE and AttrE, a lot of their time is spent on encoding the literal information. For example, in KDCoE, the time for training descriptions takes up at least $26.3\%$. By contrast, GCNAlign and MTransE use much less time, as they only use relation triples and also have a lightweight model complexity. Thus, \ul{\emph{we recognize that using auxiliary information or techniques to boost performance usually increases training time.}} Overall, MultiKE balances well between effectiveness and efficiency, as its multi-view discriminative features make it converge fast for entity alignment.} 

\section{Exploratory Experiments}
\label{sect:further_exp}

\subsection{Geometric Analysis}
\label{sect:geo_analysis}

In addition to performance comparison, we hereby focus on the geometric properties of entity embeddings, to understand how these embeddings support entity alignment performance and the underlying limitations of existing approaches.

\begin{figure}[t]
	\centering
	\includegraphics[width=\linewidth]{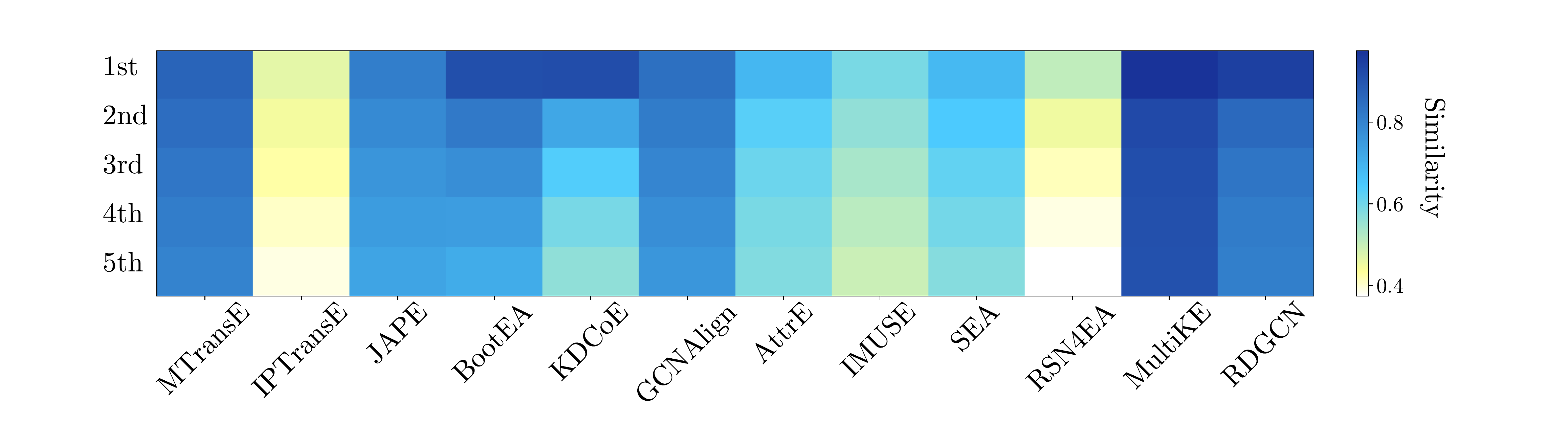}
	\centering\caption{Visualization of the similarities between entities and their top-5 nearest cross-KG neighbors on the D-Y-15K (V1) dataset. The five rows from top to bottom correspond to the similarities from the first to the fifth nearest neighbors, respectively. Darker color indicates larger similarity.}
	\label{fig:sim_map} 
\end{figure}

\subsubsection{Similarity Distribution}

Given entity embeddings, the alignment inference algorithm identifies aligned entities by the nearest neighbor search in the embedding space. It is interesting to investigate the similarity distribution of each entity and its nearest neighbors in the cross-KG scope. Towards this end, we visualize in Figure~\ref{fig:sim_map} the average similarities between entities of the source KG and their top-5 nearest neighbors of the target KG on D-Y-15K (V1). To make the similarity comparable over all the approaches, we select cosine similarity as the metric. The results show two interesting findings: 

First, the average similarities between source entities and their top-1 nearest neighbors (top-1 similarities) of different approaches differ widely. BootEA, KDCoE, MultiKE and RDGCN yield a very high top-1 similarity, while IPTransE and RSN4EA show the opposite. Intuitively, a high top-1 similarity indicates a better quality because it can reflect how confidently the entity embeddings capture the alignment information between two KGs. Most approaches with a high  top-1 similarity, such as BootEA, MultiKE and RDGCN, also achieve good performance for entity alignment (see Table \ref{tab:main_results}). For KDCoE, as shown in Figure~\ref{fig:semi_iter}, the low precision of its augmented alignment makes its top-1 entity alignment contain many errors. Hence, its performance is not as good as BootEA. But it still outperforms many others {as its description and relation embeddings are complementary, and thus can help find some correct alignment.}

Second, the similarity variances between the top-5 nearest neighbors also differ greatly, which can be reflected by the color gradients of the five rows from top to bottom. BootEA, KDCoE, RSN4EA and RDGCN exhibit large variances while MTransE, IPTransE and JAPE exhibit very slight variances. A small similarity variance means that the nearest neighbors are not discriminative enough to enable the entity to identify its counterpart correctly. {The overfitting issue in MTransE, the bootstrapping errors in IPTransE and the fuzzy entity clusters in JAPE are the reasons for their non-discriminative embeddings.} Other datasets show similar distribution. {\ul{\emph{The ideal similarity distribution for entity alignment is to hold a high top-1 similarity and a large similarity variance.}}}

\subsubsection{Hubness and Isolation}

\begin{figure}
	\centering
	\includegraphics[width=0.95\linewidth]{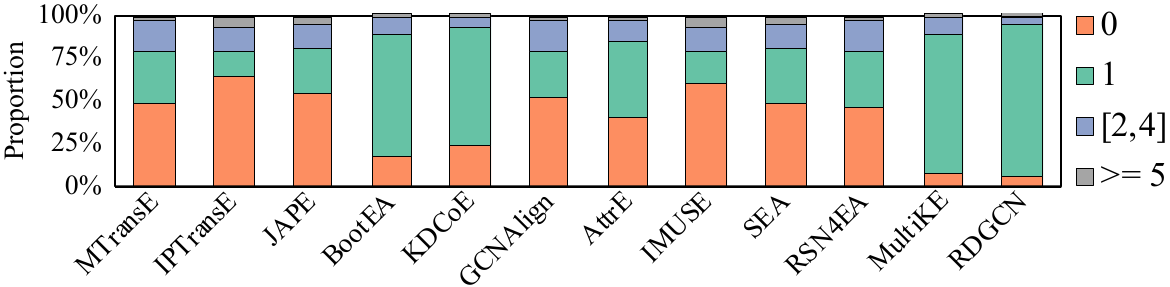}
	\centering\caption{Proportions of target entities appearing 0, 1 and more times as the nearest neighbors on D-Y-15K (V1)}
	\label{fig:csls_distribution} 
\end{figure}

Hubness is a common phenomenon in high-dimensional vector spaces \cite{Hub}, where some points (known as \emph{hubs}) frequently appear as the top-1 nearest neighbors of many other points in the vector space. Another phenomenon is that, there would exist some outliers isolated from any point clusters. The two issues have negative effects on the tasks relying on the nearest neighbor search \cite{WordTrans,WordEmbedLimit}. Here, we investigate whether embedding-based entity alignment also suffers from them. We measure the proportions of target entities that appear zero, one and more times as the nearest neighbors of source entities, respectively. Figure \ref{fig:csls_distribution} shows the results on D-Y-15K (V1), and other datasets also show similar results. Surprisingly, we find that there is a large proportion of target entities that never appear as the top-1 nearest neighbors of any source entity (marked with orange bars). This means that such isolated entities may never be considered if we use the greedy strategy of choosing the top-1 nearest neighbor to form alignment. Consequently, we would miss much correct entity alignment. The entities (blue and gray bars) that appear as the nearest neighbors of more than one source entity also occupy considerable proportions. They would cause many violations against the 1-to-1 mapping constraint and globally increase the uncertainty of alignment inference. We observe that the approaches which yield fewer isolated and hub entities, such as MultiKE and RDGCN, achieve the leading performance of entity alignment, and vice versa. So, the ideal case is to have small proportions of isolated and hub entities. {\ul{\emph{This finding indicates that we can make an estimation about the final entity alignment performance through the hubness and isolation analysis.}}}

To resolve the hubness and isolation problem, we explore cross-domain similarity local scaling (CSLS) \cite{WordTrans} as the alternative metric. It normalizes the similarity of source and target entity embeddings based on the density of their embedding neighbors. Taking cosine for example, we have
\begin{equation} 
\label{eq:csls}
\mathrm{CSLS}(\mathbf{x}_s, \mathbf{x}_t) = 2 \, \mathrm{cos}(\mathbf{x}_s, \mathbf{x}_t) - \psi_t(\mathbf{x}_s) - \psi_s(\mathbf{x}_t),
\end{equation}
where $\psi_t(\mathbf{x}_s)$ denotes the average similarity between the source entity $\mathbf{x}_s$ and its top-$k$ nearest neighbors in the target KG. $\psi_s(\mathbf{x}_t)$ is computed symmetrically. CSLS decreases the similarities between hub entities and other entities. It can also let some isolated entities be fairly considered in testing because they usually receive less similarity penalization. Therefore, we use CSLS to enhance the conventional distance metrics. In addition, we also consider stable matching (a.k.a.~stable marriage) to retrieve entity alignment from a global perspective rather than the greedy strategy based on the nearest neighbor search. The entity alignment between two KGs is stable when there does not exist another predicted aligned pair $(e_1,e_2)$ of higher preference than those of $e_1$ and $e_2$ to their current matches. The preference can be calculated based on a similarity metric such as CSLS.

\begin{table}[t]
	\centering
	\caption{Hits@1 w.r.t.~distance metrics and alignment inference strategies on D-Y-15K (V1)}
	\label{tab:csls_results}
	\begin{adjustbox}{width=\linewidth}
	\input{tab_csls}
	\end{adjustbox}
\end{table}

We report the Hits@1 results enhanced by CSLS and stable matching (abbr.~SM) in Table \ref{tab:csls_results}. We find that CSLS brings significant gains to the greedy strategy, especially on MTransE, JAPE, GCNAlign and AttrE. This is because CSLS can help mitigate the hubness phenomenon. Besides, SM further brings improvement. For example, compared with the greedy strategy, it yields a gain of more than $10\%$ on Hits@1 for MTransE, JAPE, KDCoE, GCNAlign, AttrE, IMUSE, SEA and RotatE. The reason lies in that SM can consider all entities including isolated ones. Interestingly, we observe that CSLS does not boost the performance of SM. This indicates that SM relies less on the distance metric. We gain similar results on other datasets. \ul{\emph{In summary, existing approaches concentrate on developing more powerful embedding and interaction methods, but some methods for the alignment module can also improve performance.}}

\begin{figure*}[t]
	\centering
	\setlength{\belowcaptionskip}{-5pt}
	\includegraphics[width=0.93\linewidth]{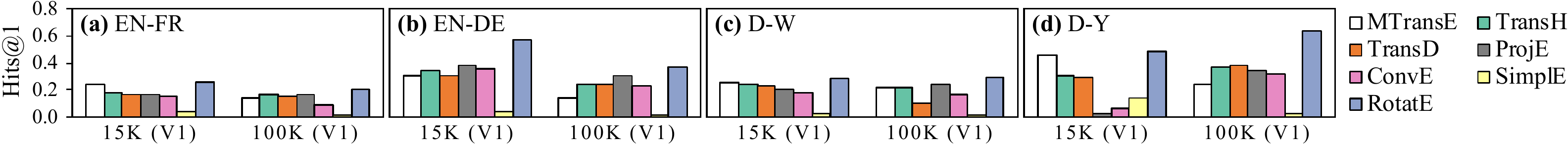}
	\centering\caption{Cross-validation results of unexplored KG embedding models on the 15K (V1) and 100K (V1) datasets}
	\label{fig:kgc_results} 
\end{figure*}

\begin{table*}[!t]
	\centering
	\caption{Comparison with conventional approaches on the 15K and 100K datasets}
	\label{tab:conventional_results}
	\begin{adjustbox}{width=\textwidth}
	\input{tab_conventional}
	\end{adjustbox}
	\vspace{-10pt}
\end{table*}

\subsection{Unexplored KG Embedding Models}
\label{sect:kgc}

As summarized in Sect.~\ref{subsect:review}, most existing approaches use TransE~\cite{TransE} or GCNs~\cite{GCN} for KG embedding due to their strong robustness and good generalizability. However, many other KG embedding models have not been explored for entity alignment yet. To fill this gap, we evaluate three translational models TransH~\cite{TransH}, TransR~\cite{TransR} and TransD~\cite{TransD}, two deep models ProjE~\cite{ProjE} and ConvE~\cite{ConvE}, as well as three semantic matching models HolE~\cite{HolE}, SimplE~\cite{SimplE} and RotatE~\cite{RotatE}, for entity alignment. We choose MTransE as baseline and replace its relation embedding model TransE with the aforementioned models. We report the Hits@1 results on the V1 datasets in Figure~\ref{fig:kgc_results}. Other results are available online. The results of TransR and HolE are omitted because their Hits@1 scores are smaller than 0.01 on most datasets. 

We can see that the improved translational models TransH and TransD show stable and promising performance on all the datasets. Specifically, on the 100K datasets, TransH is robuster than MTransE and gain better results. This is because TransH handles multi-mapping relations better and also uses negative sampling to enhance embedding. Differently, we find that TransR fails to achieve promising results. The relation-specific transformation of entity embeddings in TransR requires relation alignment to propagate the alignment information between entities. However, in our problem setting, we focus on entity alignment and do not provide relation alignment due to the great heterogeneity between KG schemata. The neural models ConvE and ProjE also show promising results on most of our datasets. However, we find that they perform poorly on D-Y-15K (V1). We owe it to the fewer relation triples and the big gap between the relation numbers in these datasets. It is difficult for the two-dimensional convolution of ConvE or the non-linear transformation of ProjE to capture the similar interactions between entity and relation embeddings across such heterogeneous KGs. For the semantic matching models, the non-Euclidean embedding model RotatE achieves much better performance than SimplE. It also outperforms other models. \ul{\emph{In short, not all KG embedding models are suitable for entity alignment, and non-Euclidean embeddings are worth further exploration.}}

\subsection{Comparison to Conventional Approaches}
\label{subsect:conventional}

We compare OpenEA with two famous open-source conventional approaches for KG alignment, i.e., LogMap \cite{LogMap} from the Semantic Web community and PARIS~\cite{PARIS} from the Database community. LogMap is an ontology matching system with built-in reasoning and inconsistency repair capabilities. PARIS is a holistic solution to align KGs based on probability estimates. The non-English KGs in cross-lingual datasets are translated to English using Google Translate to eliminate the language barrier for LogMap and PARIS. 

\noindent\textbf{Overall comparison.} Table \ref{tab:conventional_results} compares LogMap, PARIS and the best embedding-based approach in OpenEA. For the test phase of OpenEA, as each source entity gets a list of candidates, precision, recall and F1-score are in fact equal to Hits@1. All these approaches achieve good results, where PARIS performs the best on most of our datasets including EN-FR, EN-DE and D-W, and LogMap achieves promising performance on D-Y. Overall, OpenEA shows no superiority over the conventional approaches PARIS and LogMap. We think this is because current embedding-based approaches put in their main efforts in learning expressive embeddings to capture entity features while ignore the alignment inference. As summarized in Sect.~\ref{subsect:align_mod}, their alignment inference strategies are based on pairwise similarity comparison, lacking the capability of inconsistency repair and holistic estimation that LogMap and PARIS have. Our geometric analysis in Sect.~\ref{sect:geo_analysis} further shows that this weakness would lead to the issue of hubness and isolation and thus degrade entity alignment performance. By resolving this issue, as shown in Table~\ref{tab:csls_results}, OpenEA (RDGCN) achieves better Hits@1 (precision) on D-Y-15K (V1) and outperforms LogMap and PARIS in Table~\ref{tab:conventional_results}. Our experiment indicates that embedding-based entity alignment approaches require further improvement in alignment inference. Besides, we notice that LogMap fails to output entity alignment on the D-W datasets. This is because LogMap highly depends on the local names in URIs to compute similarities while the URIs in Wikidata have no actual meanings (e.g., \url{https://www.wikidata.org/wiki/Property:P69}). In fact, the results of all approaches on D-W decrease severely. The symbolic heterogeneity brings huge obstacles to both conventional and embedding-based approaches. 

\begin{table}
	\centering
	\caption{Comparison with conventional approaches using different features on EN-FR-15K (V1)}
	\label{tab:feature_results}
	\large
	\begin{adjustbox}{width=\columnwidth}
	\begin{tabular}{|l|ccc|ccc|ccc|ccc|}
	\hline \rowcolor{gray!10} & \multicolumn{3}{c|}{Using relation triples only} & \multicolumn{3}{c|}{Using attribute triples only} \\     
	\cline{2-7} \rowcolor{gray!10} & Precision & Recall & F1-score & Precision & Recall & F1-score \\ 
	\hline 
    LogMap & - & - & - & $.816_{\,\pm\, .003}$ & $.723_{\,\pm\, .002}$ & $.767_{\,\pm\, .001}$ \\ 
    PARIS & - & - & - & $.917_{\,\pm\, .000}$ & $.769_{\,\pm\, .000}$ & $.837_{\,\pm\, .000}$ \\ \hline
    BootEA & $.507_{\,\pm\, .010}$ & $.507_{\,\pm\, .010}$ & $.507_{\,\pm\, .010}$ & - & - & - \\ 
    MultiKE & $.337_{\,\pm\, .005}$ & $.337_{\,\pm\, .005}$ & $.337_{\,\pm\, .005}$ & $.719_{\,\pm\, .005}$ & $.719_{\,\pm\, .005}$ & $.719_{\,\pm\, .005}$ \\ 
    RDGCN & $.255_{\,\pm\, .004}$ & $.255_{\,\pm\, .004}$ & $.255_{\,\pm\, .004}$ & - & - & - \\ \hline
\end{tabular}
\end{adjustbox}
\end{table}

\begin{table*}
	\centering
	\caption{Summary of the required information of embedding-based and conventional entity alignment approaches}
	\label{tab:summary}
	\begin{adjustbox}{width=\textwidth}
	\begin{tabular}{|l|cccccccccccc|cc|}   
	\hline \rowcolor{gray!10} & MTransE & IPTransE & JAPE & KDCoE & BootEA & GCNAlign & AttrE & IMUSE & SEA & RSN4EA & MultiKE & RDGCN & LogMap & PARIS\\ 
	\hline 
    Relation/attribute triples & $\ast/\,~$ & $\ast/\,~$ & $\ast/\circ$ & $\circ/\circ$ & $\ast/$ & $\ast/\circ$ & $\circ/\circ$ & $\circ/\circ$ & $\ast/$ & $\ast/$ & $\circ/\circ$ & $\ast/\circ$ & $\circ/\ast$ & $\circ/\ast$\\ 
    \hline
    Pre-aligned ent./prop. & $\ast/\circ$ & $\ast/\circ$ & $\ast/\circ$ & $\ast/\,~$ & $\ast/$ & $\ast/\circ$ & $\ast/\,~$ & $\ast/\,~$ & $\ast/$ & $\ast/$ & $\ast/\circ$ & $\ast/\,~$ & $\circ/\circ$ & $\circ/\circ$\\  
    \hline    
    Word embed./Google trans. & & & & $\circ/\,~$ & & & $\circ/\,~$ & & & & $\circ/\,~$ & $\circ/\,~$ & $\,~/\triangle$ & $\,~/\triangle$ \\
    \hline \multicolumn{14}{l}{``$\ast$'' means ``mandatory'', ``$\circ$'' means ``optional'', ``$\triangle$'' means ``mandatory for cross-lingual entity alignment'', and blank means ``not applicable''.}
\end{tabular}
\end{adjustbox}
\vspace{-10pt}
\end{table*}

\noindent\textbf{Feature study.} Table~\ref{tab:feature_results} shows the results of LogMap and PARIS as well as three top-performing embedding-based approaches RDGCN, BootEA and MultiKE when only given relation or attribute triples of EN-FR-15K (V1). \ul{\emph{LogMap and PARIS rely on attribute triples and fail to output alignment in the case of using relation triples only}}. This is different from embedding-based approaches that all use relation triples. In the case of using relation triples only, BootEA is not affected by the lack of attribute triples. The performance of MultiKE and RDGCN drops greatly since their attribute embedding modules are disabled in this case. However, their relation embedding modules can still learn embeddings. When only using attribute triples, the results of LogMap almost remain intact because it mainly uses attribute triples to compute entity similarities. The recall of PARIS drops dramatically as it cannot use the relational inference to find more entity alignment. But its precision is still very high and even a little better than that in Table~\ref{tab:conventional_results}. Considering that PARIS is not designed for relational inference, relation triples may bring along noises to this approach. As for embedding-based approaches, RDGCN and BootEA cannot learn embeddings without relation triples. The multi-view approach MultiKE also suffers from performance penalties because it cannot benefit from the relation embedding. This experiment reveals the different application scenarios of these entity alignment approaches. \ul{\emph{Conventional approaches better support the entity alignment scenario with attribute information. Embedding-based approaches cover most of the typical scenarios with either relation information, attribute information or both.}}

\begin{figure}[h]
	\begin{minipage}{.38\linewidth}
	\includegraphics[width=\textwidth]{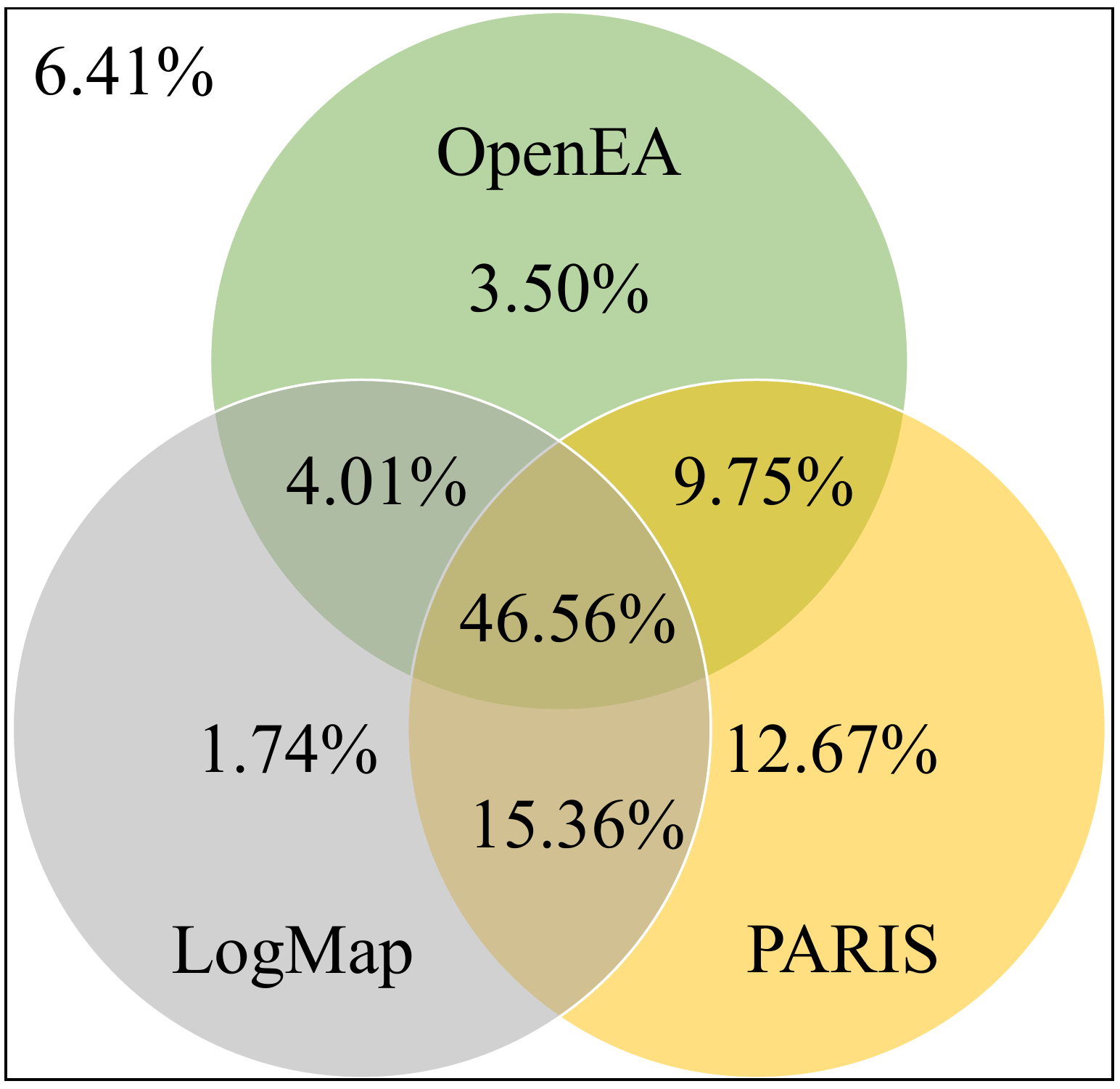}
	\end{minipage}
\hfill
	\begin{minipage}{.57\linewidth}
	\caption{\small{Proportions of the correct alignment found by LogMap, PARIS and OpenEA on EN-FR-100K (V1). OpenEA additionally finds 13.25\% (3.50\% + 9.75\%) and 7.51\% (3.50\% + 4.01\%) of the alignment that LogMap and PARIS do not, respectively. Besides, 6.41\% of the alignment is not found by any approach, while 45.56\% is found by all the three.}}
	\label{fig:pie} 
	\end{minipage}
\end{figure}

\noindent\textbf{Analysis on predicted alignment.} To further investigate the potential complementarity of embedding-based and conventional approaches, we show in Figure \ref{fig:pie} the proportions of correct alignment found by OpenEA (RDGCN), LogMap and PARIS on EN-FR-100K (V1). They all suffer from the same challenge (the symbolic heterogeneity). We find that they can produce complementary entity alignment. This analysis calls for a hybrid system for entity alignment built on both conventional and embedding-based techniques.

\section{Summary and Future Directions}
\label{sect:future}

\subsection{Summary of Experiments}
From our experimental results, we find that (i) RDGCN, BootEA and MultiKE achieve the most competitive performance. This suggests that incorporating both literal information and carefully-designed bootstrapping can help entity alignment. (ii) For the embedding models designed for link prediction, we find that not all of them are suitable for entity alignment. (iii) Currently, the alignment inference strategy receives little attention. Our preliminary results show that the CSLS distance metric and the stable matching strategy can bring performance improvement to all the approaches. (iv) We also find that embedding-based and conventional entity alignment approaches are complementary to each other. (v) For choosing appropriate approaches based on the available resources in real-world scenarios, Table~\ref{tab:summary} summarizes the required information of embedding-based and conventional entity alignment approaches in our experimental analysis.

\subsection{Future Directions}

\noindent\textbf{Unsupervised entity alignment.} As summarized in Section~\ref{subsect:interact} and discussed in Section \ref{subsect:main}, all current approaches require seed alignment as supervision. However, this requirement is sometimes difficult to be satisfied in the real world. Hence, studying unsupervised entity alignment is a meaningful direction. A possible solution is to incorporate auxiliary features or resources and distill distant supervision from them, such as discriminative features (homepages of people and introductory images of products) and pre-trained word embeddings \cite{HMAN}. Besides, recent advances in unsupervised cross-lingual word alignment \cite{WordTrans} like orthogonal Procrustes \cite{Procrustes} and adversarial training \cite{GAN} are also worth investigation. Another possible solution is to use active learning \cite{GenLink,ActiveER} or abductive learning \cite{Abductive_learning} to reduce the burden of data labeling.

\noindent\textbf{Long-tail entity alignment.} Our experimental analysis on the sparse and dense datasets reveals the difficulty in aligning long-tail entities, which usually account for a large proportion in KGs \cite{LongTail}. To embed long-tail entities, in addition to using more advanced graph neural networks \cite{GCN,RGCN,GAT}, injecting more features such as multi-modal data and taxonomies would also be helpful. As KGs are far from complete, jointly training link prediction and entity alignment via a unified framework may leverage the incidental supervision of both tasks. Extracting additional information from the open Web to enrich long-tail entities is also a potential direction~\cite{ConMask}.

\noindent\textbf{Large-scale entity alignment.} The running time comparison shows that training existing approaches on larger datasets costs much more time. The test phrase also takes much time. For example, computing the pairwise cosine similarity of entity embeddings on a 100K dataset costs about 8 minutes by using 10 processes in parallel. The cost would grow polynomially along with the growing number of entities. It is difficult for embedding-based (and also conventional) approaches to run on very large KGs due to the large and unpartitioned candidate space. The \emph{blocking} techniques, e.g., locality-sensitive hashing~\cite{LSH} and hashing representation learning \cite{Hashing}, may be useful to narrow the candidate space.

\noindent\textbf{Entity alignment in non-Euclidean spaces.} Our experimental results in Figure \ref{fig:kgc_results} indicate that the non-Euclidean embedding model RotatE \cite{RotatE} outperforms other Euclidean models. We also notice that recent non-Euclidean embeddings have demonstrated their effectiveness in representing graph-structured data \cite{Hyperbolic}. So, alignment-oriented non-Euclidean KG embedding models are worth exploiting.

\section{Conclusion}
\label{sect:concl}

In this paper, we survey the field of embedding-based entity alignment between KGs and conduct a benchmarking study of the representative approaches. We create a set of dedicated datasets that better fit real-world KGs and develop an open-source library containing a variety of entity alignment approaches and KG embedding models. Our experiments analyze the status quo and point out future directions. 

\smallskip\noindent\textbf{Acknowledgments.} 
This work was supported by the National Key R\&D Program of China (No. 2018YFB1004300), the National Natural Science Foundation of China (No. 61872172), and the Collaborative Innovation Center of Novel Software Technology and Industrialization.

\bibliographystyle{abbrv}
\bibliography{vldb_ref}  

\end{document}

%% file: tab_categ.tex
\begin{tabular}{|l|c|c|c|c|c|}
	\hline \rowcolor{gray!10} & \multicolumn{2}{c|}{Embedding} & Alignment & \multicolumn{2}{c|}{Interaction} \\
	\cline{2-6} \cellcolor{gray!10} \multirow{-2}{*}{} & Relation & Att. & Emb. distance & Combination & Learning \\ 
	\hline	
	        MTransE \cite{MTransE} & Triple & - & Euclidean & Transformation & Superv. \\ 	 
		 	IPTransE \cite{IPTransE} & Path & - & Euclidean & Sharing & Semi- \\ 
		 	JAPE \cite{JAPE} & Triple & Att. & Cosine & Sharing & Superv. \\  
		 	BootEA \cite{BootEA} & Triple & - & Cosine & Swapping & Semi- \\  
		 	KDCoE \cite{KDCoE} & Triple & Literal & Euclidean & Transformation & Semi- \\ 
		 	NTAM~\cite{NTAM} & Triple & - & Cosine & Swapping & Superv. \\ 
		 	GCNAlign \cite{GCNAlign} & Neighbor & Att. & Manhattan & Calibration & Superv. \\ 
		 	AttrE \cite{AttrE} & Triple & Literal & Cosine & Sharing & Superv. \\ 
		 	IMUSE \cite{IMUSE} & Triple & Literal & Cosine & Sharing & Superv.\\
		 	SEA \cite{SEA} & Triple & - & Cosine & Transformation &  Superv. \\
		 	RSN4EA \cite{RSN} & Path & - & Cosine & Sharing & Superv. \\
		 	GMNN \cite{GMNN} & Neighbor & Literal & Cosine & Swapping & Superv. \\
		 	MuGNN \cite{MuGNN} & Neighbor & - & Manhattan & Calibration & Superv. \\
		 	OTEA \cite{OTEA} & Triple & - & Euclidean & Transformation & Superv.\\
		    NAEA \cite{NAEA} & Neighbor & - & Cosine & Swapping & Superv.\\
		    AVR-GCN \cite{AVR-GCN} & Neighbor & - & Euclidean & Swapping & Superv. \\
		 	MultiKE \cite{MultiKE} & Triple & Literal & Cosine & Swapping & Superv. \\
		 	RDGCN \cite{RDGCN} & Neighbor & Literal & Manhattan & Calibration & Superv. \\
		 	KECG \cite{KECG} & Neighbor & - & Euclidean & Calibration & Superv. \\
		 	HGCN \cite{HGCN} & Neighbor & Literal & Euclidean & Calibration & Superv. \\
			MMEA \cite{MMEA} & Triple & - & Cosine & Sharing & Superv. \\
			HMAN \cite{HMAN} & Neighbor & Literal & Euclidean & Calibration & Superv. \\
			AKE \cite{AKE} & Triple & - & Euclidean & Transformation & Superv. \\
	\hline  
\end{tabular}

%% file: tab_stats.tex
\begin{tabular}{|cc|rrrr|rrrr|rrrr|rrrr|}
\hline \rowcolor{gray!10} & & \multicolumn{4}{c|}{15K (V1)} & \multicolumn{4}{c|}{15K (V2)} & \multicolumn{4}{c|}{100K (V1)} & \multicolumn{4}{c|}{100K (V2)} \\
\cline{3-18} \cellcolor{gray!10} \multirow{-2}{*}{Datasets} & \cellcolor{gray!10} \multirow{-2}{*}{KGs} 
	& \#Rel. & \#Att. & \#Rel tr. & \#Att tr. & \#Rel. & \#Att. &  \#Rel tr. & \#Att tr. 
	& \#Rel. & \#Att. & \#Rel tr. & \#Att tr. & \#Rel. & \#Att. & \#Rel tr. & \#Att tr. \\ 
\hline \multirow{2}{*}{EN-FR}  
	& EN & 267 & 308 & 47,334 & 73,121 & 193 & 189 & 96,318 & 66,899 & 400 & 466 & 309,607 & 497,729 & 379 & 364 & 649,902 & 503,922 \\
	& FR & 210 & 404 & 40,864 & 67,167 & 166 & 221 & 80,112 & 68,779 & 300 & 519 & 258,285 & 426,672 & 287 & 468 & 561,391 & 431,379 \\
\hline \multirow{2}{*}{EN-DE}  
	& EN & 215 & 286 & 47,676 & 83,755 & 169 & 171 & 84,867 & 81,988 & 381 & 451 & 335,359 & 552,750 & 323 & 326 & 622,588 & 560,247 \\
	& DE & 131 & 194 & 50,419 & 156,150 & 96 & 116 & 92,632 & 186,335 & 196 & 252 & 336,240 & 716,615 & 170 & 189 & 629,395 & 793,710 \\
\hline \multirow{2}{*}{D-W}  
	& DB & 248 & 342 & 38,265 & 68,258 & 167 & 175 & 73,983 & 66,813 & 413 & 493 & 293,990 & 451,011 & 318 & 328 & 616,457 & 467,103 \\
	& WD & 169 & 649 & 42,746 & 138,246 & 121 & 457 & 83,365 & 175,686 & 261 & 874 & 251,708 & 687,860 & 239 & 760 & 588,203 & 878,219 \\
\hline \multirow{2}{*}{D-Y}  
	& DB & 165 & 257 & 30,291 & 71,716 & 72 & 90 & 68,063 & 65,100 & 287 & 379 & 294,188 & 523,062 & 230 & 277 & 576,547 & 547,026 \\
	& YG & 28 & 35 & 26,638 & 132,114 & 21 & 20 & 60,970 & 131,151 & 32 & 38 & 400,518 & 749,787 & 31 & 36 & 865,265 & 855,161 \\
\hline
\end{tabular}

%% file: tab_main_results.tex
\begin{tabular}{|l|l|ccc|ccc|ccc|ccc|} 
\hline \rowcolor{gray!10} \multicolumn{2}{|c|}{} & \multicolumn{3}{c|}{15K (V1)} & \multicolumn{3}{c|}{15K (V2)} & \multicolumn{3}{c|}{100K (V1)} & \multicolumn{3}{c|}{100K (V2)} \\     
\cline{3-14} \rowcolor{gray!10} \multicolumn{2}{|c|}{} & Hits@1 & Hits@5 & MRR & Hits@1 & Hits@5 & MRR & Hits@1 & Hits@5 & MRR & Hits@1 & Hits@5 & MRR \\
\hline \parbox[t]{2mm}{\multirow{11}{*}{\rotatebox[origin=c]{90}{EN-FR}}}
& MTransE	& $.247_{\,\pm\, .006}$& $.467_{\,\pm\, .009}$& $.351_{\,\pm\, .007}$
   			& $.240_{\,\pm\, .005}$& $.436_{\,\pm\, .007}$& $.336_{\,\pm\, .005}$
    			& $.138_{\,\pm\, .002}$& $.261_{\,\pm\, .004}$& $.202_{\,\pm\, .002}$
    			& $.090_{\,\pm\, .003}$& $.174_{\,\pm\, .003}$& $.135_{\,\pm\, .003}$
\\
& IPTransE	& $.169_{\,\pm\, .013}$& $.320_{\,\pm\, .025}$& $.243_{\,\pm\, .019}$
    			& $.236_{\,\pm\, .012}$& $.449_{\,\pm\, .021}$& $.339_{\,\pm\, .016}$
   			& $.158_{\,\pm\, .004}$& $.277_{\,\pm\, .008}$& $.219_{\,\pm\, .006}$
    			& $.234_{\,\pm\, .007}$& $.431_{\,\pm\, .015}$& $.329_{\,\pm\, .010}$
\\
& JAPE	& $.262_{\,\pm\, .006}$& $.497_{\,\pm\, .010}$& $.372_{\,\pm\, .007}$
    		& $.292_{\,\pm\, .009}$& $.524_{\,\pm\, .006}$& $.402_{\,\pm\, .007}$
    		& $.165_{\,\pm\, .002}$& $.310_{\,\pm\, .002}$& $.240_{\,\pm\, .002}$
    		& $.125_{\,\pm\, .003}$& $.239_{\,\pm\, .005}$& $.183_{\,\pm\, .004}$
\\
& KDCoE	& \textcolor{cyan}{$.581_{\,\pm\, .004}$}& $.680_{\,\pm\, .004}$& $.628_{\,\pm\, .003}$
    		& \textcolor{cyan}{$.730_{\,\pm\, .007}$}& $.837_{\,\pm\, .006}$& \textcolor{cyan}{$.778_{\,\pm\, .005}$}
    		& \textcolor{cyan}{$.482_{\,\pm\, .005}$}& $.515_{\,\pm\, .006}$& \textcolor{cyan}{$.499_{\,\pm\, .005}$}
    		& $.611_{\,\pm\, .012}$& $.653_{\,\pm\, .015}$& $.632_{\,\pm\, .014}$
\\
& BootEA	& $.507_{\,\pm\, .010}$& \textcolor{cyan}{$.718_{\,\pm\, .012}$}& $.603_{\,\pm\, .011}$
    		& $.660_{\,\pm\, .006}$& \textcolor{cyan}{$.850_{\,\pm\, .005}$}& $.745_{\,\pm\, .005}$
    		& $.389_{\,\pm\, .004}$& $.561_{\,\pm\, .004}$& $.474_{\,\pm\, .004}$
    		& \textcolor{cyan}{$.640_{\,\pm\, .001}$}& \textcolor{red}{$.806_{\,\pm\, .001}$}& \textcolor{blue}{$.716_{\,\pm\, .000}$}
\\
& GCNAlign	& $.338_{\,\pm\, .002}$& $.589_{\,\pm\, .009}$& $.451_{\,\pm\, .005}$
    			& $.414_{\,\pm\, .005}$& $.698_{\,\pm\, .007}$& $.542_{\,\pm\, .005}$
   		 	& $.230_{\,\pm\, .002}$& $.412_{\,\pm\, .004}$& $.319_{\,\pm\, .003}$
    			& $.257_{\,\pm\, .002}$& $.455_{\,\pm\, .003}$& $.351_{\,\pm\, .002}$
\\
& AttrE	& $.481_{\,\pm\, .010}$& $.671_{\,\pm\, .009}$& $.569_{\,\pm\, .010}$
    		& $.535_{\,\pm\, .015}$& $.746_{\,\pm\, .014}$& $.631_{\,\pm\, .014}$
    		& $.403_{\,\pm\, .019}$& \textcolor{cyan}{$.572_{\,\pm\, .019}$}& $.483_{\,\pm\, .019}$
   		& $.466_{\,\pm\, .011}$& $.644_{\,\pm\, .012}$& $.549_{\,\pm\, .011}$
\\
& IMUSE	& $.569_{\,\pm\, .006}$& $.717_{\,\pm\, .010}$& \textcolor{cyan}{$.638_{\,\pm\, .008}$}
    		& $.607_{\,\pm\, .013}$& $.760_{\,\pm\, .014}$& $.678_{\,\pm\, .013}$
    		& $.439_{\,\pm\, .002}$& $.546_{\,\pm\, .004}$& $.492_{\,\pm\, .003}$
    		& $.461_{\,\pm\, .003}$& $.605_{\,\pm\, .005}$& $.529_{\,\pm\, .004}$
\\
& SEA	& $.280_{\,\pm\, .015}$& $.530_{\,\pm\, .026}$& $.397_{\,\pm\, .019}$
    		& $.360_{\,\pm\, .018}$& $.651_{\,\pm\, .018}$& $.494_{\,\pm\, .017}$
   		& $.225_{\,\pm\, .011}$& $.399_{\,\pm\, .013}$& $.314_{\,\pm\, .012}$
    		& $.297_{\,\pm\, .002}$& $.500_{\,\pm\, .002}$& $.395_{\,\pm\, .002}$
\\
& RSN4EA	& $.393_{\,\pm\, .007}$& $.595_{\,\pm\, .012}$& $.487_{\,\pm\, .009}$
    			& $.579_{\,\pm\, .006}$& $.759_{\,\pm\, .006}$& $.662_{\,\pm\, .006}$
    			& $.293_{\,\pm\, .004}$& $.452_{\,\pm\, .006}$& $.371_{\,\pm\, .004}$
    			& $.495_{\,\pm\, .003}$& $.672_{\,\pm\, .005}$& $.578_{\,\pm\, .004}$
	
\\
& MultiKE	& \textcolor{blue}{$.749_{\,\pm\, .004}$}& \textcolor{blue}{$.819_{\,\pm\, .005}$}& \textcolor{blue}{$.782_{\,\pm\, .004}$}
    		& \textcolor{red}{$.864_{\,\pm\, .007}$}& \textcolor{blue}{$.909_{\,\pm\, .005}$}& \textcolor{red}{$.885_{\,\pm\, .006}$}
    		& \textcolor{blue}{$.629_{\,\pm\, .002}$}& \textcolor{blue}{$.680_{\,\pm\, .002}$}& \textcolor{blue}{$.655_{\,\pm\, .002}$}
    		& \textcolor{blue}{$.642_{\,\pm\, .003}$}& \textcolor{cyan}{$.696_{\,\pm\, .003}$}& \textcolor{cyan}{$.670_{\,\pm\, .003}$}
\\
& RDGCN	& \textcolor{red}{$.755_{\,\pm\, .004}$}& \textcolor{red}{$.854_{\,\pm\, .003}$}& \textcolor{red}{$.800_{\,\pm\, .003}$}
    		& \textcolor{blue}{$.847_{\,\pm\, .006}$}& \textcolor{red}{$.919_{\,\pm\, .004}$}& \textcolor{blue}{$.880_{\,\pm\, .005}$}
    		& \textcolor{red}{$.640_{\,\pm\, .004}$}& \textcolor{red}{$.732_{\,\pm\, .004}$}& \textcolor{red}{$.683_{\,\pm\, .004}$}
    		& \textcolor{red}{$.715_{\,\pm\, .003}$}& \textcolor{blue}{$.787_{\,\pm\, .002}$}& \textcolor{red}{$.748_{\,\pm\, .002}$}
\\
\hline \parbox[t]{2mm}{\multirow{11}{*}{\rotatebox[origin=c]{90}{EN-DE}}}
& MTransE	& $.307_{\,\pm\, .007}$& $.518_{\,\pm\, .004}$& $.407_{\,\pm\, .006}$
    			& $.193_{\,\pm\, .016}$& $.352_{\,\pm\, .023}$& $.274_{\,\pm\, .018}$
    			& $.140_{\,\pm\, .003}$& $.264_{\,\pm\, .004}$& $.204_{\,\pm\, .004}$
    			& $.115_{\,\pm\, .003}$& $.215_{\,\pm\, .004}$& $.168_{\,\pm\, .003}$
\\
& IPTransE	& $.350_{\,\pm\, .009}$& $.515_{\,\pm\, .012}$& $.43_{\,\pm\, .011}$
    			& $.476_{\,\pm\, .012}$& $.678_{\,\pm\, .011}$& $.571_{\,\pm\, .010}$
    			& $.226_{\,\pm\, .014}$& $.357_{\,\pm\, .019}$& $.292_{\,\pm\, .017}$
    			& $.346_{\,\pm\, .013}$& $.535_{\,\pm\, .016}$& $.437_{\,\pm\, .014}$
\\
& JAPE	& $.288_{\,\pm\, .016}$& $.512_{\,\pm\, .018}$& $.394_{\,\pm\, .016}$
    		& $.167_{\,\pm\, .011}$& $.329_{\,\pm\, .015}$& $.250_{\,\pm\, .013}$
    		& $.152_{\,\pm\, .006}$& $.291_{\,\pm\, .009}$& $.223_{\,\pm\, .007}$
    		& $.11_{\,\pm\, .004}$& $.218_{\,\pm\, .006}$& $.167_{\,\pm\, .005}$
\\
& KDCoE	& $.529_{\,\pm\, .014}$& $.629_{\,\pm\, .015}$& $.580_{\,\pm\, .014}$
    		& $.649_{\,\pm\, .017}$& $.788_{\,\pm\, .017}$& $.715_{\,\pm\, .016}$
    		& $.506_{\,\pm\, .014}$& $.591_{\,\pm\, .019}$& $.549_{\,\pm\, .016}$
    		& $.651_{\,\pm\, .011}$& $.756_{\,\pm\, .010}$& \textcolor{cyan}{$.701_{\,\pm\, .011}$} 
\\
& BootEA	& \textcolor{cyan}{$.675_{\,\pm\, .004}$}& \textcolor{blue}{$.820_{\,\pm\, .004}$}& \textcolor{cyan}{$.740_{\,\pm\, .004}$}
    		& \textcolor{red}{$.833_{\,\pm\, .015}$}& \textcolor{red}{$.912_{\,\pm\, .008}$}& \textcolor{red}{$.869_{\,\pm\, .012}$}
    		& \textcolor{cyan}{$.518_{\,\pm\, .003}$}& \textcolor{cyan}{$.673_{\,\pm\, .003}$}& \textcolor{cyan}{$.592_{\,\pm\, .003}$}
    		& \textcolor{blue}{$.739_{\,\pm\, .004}$}& \textcolor{red}{$.851_{\,\pm\, .003}$}& \textcolor{blue}{$.791_{\,\pm\, .004}$}
\\
& GCNAlign	& $.481_{\,\pm\, .003}$& $.679_{\,\pm\, .005}$& $.571_{\,\pm\, .003}$
    			& $.534_{\,\pm\, .005}$& $.717_{\,\pm\, .005}$& $.618_{\,\pm\, .005}$
    			& $.317_{\,\pm\, .007}$& $.485_{\,\pm\, .008}$& $.399_{\,\pm\, .007}$
    			& $.375_{\,\pm\, .005}$& $.549_{\,\pm\, .006}$& $.457_{\,\pm\, .005}$
\\
& AttrE	& $.517_{\,\pm\, .011}$& $.687_{\,\pm\, .013}$& $.597_{\,\pm\, .011}$
    		& $.650_{\,\pm\, .015}$& $.816_{\,\pm\, .008}$& $.726_{\,\pm\, .012}$
    		& $.399_{\,\pm\, .010}$& $.554_{\,\pm\, .012}$& $.473_{\,\pm\, .011}$
    		& $.464_{\,\pm\, .011}$& $.637_{\,\pm\, .010}$& $.546_{\,\pm\, .011}$
\\
& IMUSE	& $.580_{\,\pm\, .017}$& $.720_{\,\pm\, .014}$& $.647_{\,\pm\, .015}$
    		& $.674_{\,\pm\, .011}$& $.803_{\,\pm\, .008}$& $.734_{\,\pm\, .010}$
    		& $.421_{\,\pm\, .005}$& $.516_{\,\pm\, .005}$& $.469_{\,\pm\, .005}$
    		& $.457_{\,\pm\, .005}$& $.588_{\,\pm\, .007}$& $.521_{\,\pm\, .006}$

\\
& SEA	& $.530_{\,\pm\, .027}$& $.718_{\,\pm\, .026}$& $.617_{\,\pm\, .025}$
    		& $.606_{\,\pm\, .024}$& $.779_{\,\pm\, .018}$& $.687_{\,\pm\, .020}$
    		& $.341_{\,\pm\, .016}$& $.502_{\,\pm\, .017}$& $.421_{\,\pm\, .016}$
    		& $.447_{\,\pm\, .006}$& $.625_{\,\pm\, .006}$& $.532_{\,\pm\, .006}$
\\
& RSN4EA	& $.587_{\,\pm\, .001}$& $.752_{\,\pm\, .003}$& $.662_{\,\pm\, .001}$
    			& \textcolor{cyan}{$.791_{\,\pm\, .009}$}& \textcolor{cyan}{$.890_{\,\pm\, .006}$}& \textcolor{cyan}{$.837_{\,\pm\, .008}$}
    			& $.430_{\,\pm\, .002}$& $.57_{\,\pm\, .001}$& $.497_{\,\pm\, .001}$
    			& $.639_{\,\pm\, .001}$& \textcolor{cyan}{$.763_{\,\pm\, .001}$}& $.697_{\,\pm\, .001}$
\\
& MultiKE	& \textcolor{blue}{$.756_{\,\pm\, .004}$}& \textcolor{cyan}{$.809_{\,\pm\, .003}$}& \textcolor{blue}{$.782_{\,\pm\, .003}$}
    		& $.755_{\,\pm\, .008}$& $.813_{\,\pm\, .008}$& $.784_{\,\pm\, .007}$
    		& \textcolor{blue}{$.668_{\,\pm\, .002}$}& \textcolor{blue}{$.712_{\,\pm\, .002}$}& \textcolor{blue}{$.690_{\,\pm\, .001}$}
    		& \textcolor{cyan}{$.661_{\,\pm\, .004}$}& $.709_{\,\pm\, .004}$& $.686_{\,\pm\, .004}$
\\
& RDGCN	& \textcolor{red}{$.830_{\,\pm\, .006}$}& \textcolor{red}{$.895_{\,\pm\, .004}$}&                          \textcolor{red}{$.859_{\,\pm\, .005}$}
    		& \textcolor{blue}{$.833_{\,\pm\, .007}$}& \textcolor{blue}{$.891_{\,\pm\, .005}$}& \textcolor{blue}{$.860_{\,\pm\, .006}$}
    		& \textcolor{red}{$.722_{\,\pm\, .002}$}& \textcolor{red}{$.794_{\,\pm\, .002}$}& \textcolor{red}{$.756_{\,\pm\, .002}$}
    		& \textcolor{red}{$.766_{\,\pm\, .002}$}& \textcolor{blue}{$.829_{\,\pm\, .002}$}& \textcolor{red}{$.796_{\,\pm\, .002}$}
\\
\hline \parbox[t]{2mm}{\multirow{11}{*}{\rotatebox[origin=c]{90}{D-W}}}
& MTransE	& $.259_{\,\pm\, .008}$& $.461_{\,\pm\, .012}$& $.354_{\,\pm\, .008}$
    			& $.271_{\,\pm\, .013}$& $.49_{\,\pm\, .014}$& $.376_{\,\pm\, .013}$
    			& $.210_{\,\pm\, .003}$& $.358_{\,\pm\, .003}$& $.282_{\,\pm\, .003}$
    			& $.148_{\,\pm\, .004}$& $.268_{\,\pm\, .005}$& $.209_{\,\pm\, .005}$
\\
& IPTransE	& $.232_{\,\pm\, .012}$& $.38_{\,\pm\, .016}$& $.303_{\,\pm\, .014}$
    			& $.412_{\,\pm\, .007}$& $.623_{\,\pm\, .010}$& $.511_{\,\pm\, .007}$
    			& $.221_{\,\pm\, .004}$& $.352_{\,\pm\, .008}$& $.285_{\,\pm\, .006}$
    			& $.319_{\,\pm\, .017}$& $.516_{\,\pm\, .024}$& $.413_{\,\pm\, .020}$
\\
& JAPE	& $.250_{\,\pm\, .007}$& $.457_{\,\pm\, .010}$& $.348_{\,\pm\, .007}$
    		& $.262_{\,\pm\, .013}$& $.484_{\,\pm\, .019}$& $.368_{\,\pm\, .015}$
    		& $.211_{\,\pm\, .004}$& $.369_{\,\pm\, .004}$& $.287_{\,\pm\, .004}$
    		& $.154_{\,\pm\, .004}$& $.287_{\,\pm\, .005}$& $.221_{\,\pm\, .005}$
\\
& KDCoE	& $.247_{\,\pm\, .020}$& $.412_{\,\pm\, .029}$& $.325_{\,\pm\, .023}$
    		& $.405_{\,\pm\, .020}$& $.640_{\,\pm\, .019}$& $.515_{\,\pm\, .020}$
    		& $.157_{\,\pm\, .003}$& $.243_{\,\pm\, .007}$& $.199_{\,\pm\, .005}$
    		& $.373_{\,\pm\, .010}$& $.550_{\,\pm\, .014}$& $.458_{\,\pm\, .012}$
\\
& BootEA	& \textcolor{red}{$.572_{\,\pm\, .008}$}& \textcolor{red}{$.744_{\,\pm\, .007}$}& \textcolor{red}{$.649_{\,\pm\, .008}$}
    		& \textcolor{red}{$.821_{\,\pm\, .004}$}& \textcolor{red}{$.926_{\,\pm\, .003}$}& \textcolor{red}{$.867_{\,\pm\, .003}$}
    		& \textcolor{red}{$.516_{\,\pm\, .006}$}& \textcolor{red}{$.685_{\,\pm\, .006}$}& \textcolor{red}{$.594_{\,\pm\, .005}$}
    		& \textcolor{red}{$.766_{\,\pm\, .007}$}& \textcolor{red}{$.892_{\,\pm\, .005}$}& \textcolor{red}{$.822_{\,\pm\, .006}$}
\\
& GCNAlign	& $.364_{\,\pm\, .009}$& $.580_{\,\pm\, .010}$& $.461_{\,\pm\, .008}$
    			& $.506_{\,\pm\, .006}$& $.743_{\,\pm\, .005}$& $.612_{\,\pm\, .005}$
    			& $.324_{\,\pm\, .002}$& \textcolor{cyan}{$.507_{\,\pm\, .004}$}& $.409_{\,\pm\, .003}$
    			& $.353_{\,\pm\, .004}$& $.559_{\,\pm\, .006}$& $.449_{\,\pm\, .004}$
\\
& AttrE	& $.299_{\,\pm\, .004}$& $.467_{\,\pm\, .003}$& $.381_{\,\pm\, .003}$
    		& $.489_{\,\pm\, .016}$& $.695_{\,\pm\, .016}$& $.585_{\,\pm\, .015}$
    		& $.209_{\,\pm\, .008}$& $.335_{\,\pm\, .011}$& $.273_{\,\pm\, .009}$
    		& $.301_{\,\pm\, .015}$& $.475_{\,\pm\, .018}$& $.386_{\,\pm\, .016}$
\\
& IMUSE	& $.327_{\,\pm\, .016}$& $.523_{\,\pm\, .024}$& $.419_{\,\pm\, .019}$
    		& $.581_{\,\pm\, .016}$& \textcolor{cyan}{$.778_{\,\pm\, .011}$}& $.671_{\,\pm\, .014}$
    		& $.276_{\,\pm\, .010}$& $.437_{\,\pm\, .016}$& $.355_{\,\pm\, .013}$
    		& \textcolor{cyan}{$.431_{\,\pm\, .011}$}& \textcolor{cyan}{$.631_{\,\pm\, .013}$}& \textcolor{cyan}{$.525_{\,\pm\, .012}$}
\\
& SEA	& $.360_{\,\pm\, .012}$& $.572_{\,\pm\, .015}$& $.458_{\,\pm\, .013}$
    		& $.567_{\,\pm\, .008}$& $.770_{\,\pm\, .007}$& $.660_{\,\pm\, .008}$
    		& $.291_{\,\pm\, .012}$& $.470_{\,\pm\, .014}$& $.378_{\,\pm\, .013}$
    		& $.382_{\,\pm\, .003}$& $.585_{\,\pm\, .003}$& $.479_{\,\pm\, .002}$
\\
& RSN4EA	& \textcolor{cyan}{$.441_{\,\pm\, .008}$}& \textcolor{cyan}{$.615_{\,\pm\, .007}$}& \textcolor{cyan}{$.521_{\,\pm\, .007}$}
    			& \textcolor{blue}{$.723_{\,\pm\, .007}$}& \textcolor{blue}{$.854_{\,\pm\, .006}$}& \textcolor{blue}{$.782_{\,\pm\, .006}$}
    			& \textcolor{blue}{$.384_{\,\pm\, .004}$}& \textcolor{blue}{$.533_{\,\pm\, .006}$}& \textcolor{blue}{$.454_{\,\pm\, .005}$}
    			& \textcolor{blue}{$.634_{\,\pm\, .004}$}& \textcolor{blue}{$.776_{\,\pm\, .002}$}& \textcolor{blue}{$.699_{\,\pm\, .003}$}
\\
& MultiKE	& $.411_{\,\pm\, .010}$& $.521_{\,\pm\, .017}$& $.468_{\,\pm\, .012}$
    		& $.495_{\,\pm\, .010}$& $.646_{\,\pm\, .016}$& $.569_{\,\pm\, .011}$
    		& $.290_{\,\pm\, .006}$& $.357_{\,\pm\, .011}$& $.326_{\,\pm\, .009}$
    		& $.319_{\,\pm\, .001}$& $.396_{\,\pm\, .004}$& $.360_{\,\pm\, .002}$
\\
& RDGCN	& \textcolor{blue}{$.515_{\,\pm\, .008}$}& \textcolor{blue}{$.669_{\,\pm\, .006}$}& \textcolor{blue}{$.584_{\,\pm\, .007}$}
    		& \textcolor{cyan}{$.623_{\,\pm\, .006}$}& $.757_{\,\pm\, .004}$& \textcolor{cyan}{$.684_{\,\pm\, .005}$}
    		& \textcolor{cyan}{$.362_{\,\pm\, .002}$}& $.485_{\,\pm\, .002}$& \textcolor{cyan}{$.420_{\,\pm\, .002}$}
    		& $.421_{\,\pm\, .001}$& $.528_{\,\pm\, .002}$& $.473_{\,\pm\, .002}$
\\
\hline \parbox[t]{2mm}{\multirow{11}{*}{\rotatebox[origin=c]{90}{D-Y}}}
& MTransE	& $.463_{\,\pm\, .013}$& $.675_{\,\pm\, .011}$& $.559_{\,\pm\, .012}$
    			& $.443_{\,\pm\, .017}$& $.635_{\,\pm\, .013}$& $.533_{\,\pm\, .015}$
    			& $.244_{\,\pm\, .004}$& $.414_{\,\pm\, .006}$& $.328_{\,\pm\, .005}$
    			& $.100_{\,\pm\, .003}$& $.195_{\,\pm\, .005}$& $.152_{\,\pm\, .004}$
\\
& IPTransE	& $.313_{\,\pm\, .009}$& $.456_{\,\pm\, .015}$& $.378_{\,\pm\, .011}$
    			& $.752_{\,\pm\, .018}$& $.873_{\,\pm\, .013}$& $.808_{\,\pm\, .015}$
    			& $.396_{\,\pm\, .014}$& $.558_{\,\pm\, .018}$& $.474_{\,\pm\, .015}$
    			& $.456_{\,\pm\, .016}$& $.620_{\,\pm\, .017}$& $.534_{\,\pm\, .017}$
\\
& JAPE	& $.469_{\,\pm\, .009}$& $.687_{\,\pm\, .011}$& $.567_{\,\pm\, .009}$
    		& $.345_{\,\pm\, .010}$& $.546_{\,\pm\, .013}$& $.440_{\,\pm\, .011}$
    		& $.287_{\,\pm\, .007}$& $.474_{\,\pm\, .008}$& $.379_{\,\pm\, .007}$
    		& $.127_{\,\pm\, .004}$& $.244_{\,\pm\, .006}$& $.189_{\,\pm\, .005}$
\\
& KDCoE	& $.661_{\,\pm\, .013}$& $.764_{\,\pm\, .036}$& $.710_{\,\pm\, .021}$
    		& $.895_{\,\pm\, .013}$& \textcolor{blue}{$.974_{\,\pm\, .003}$}& $.932_{\,\pm\, .008}$
    		& $.565_{\,\pm\, .001}$& $.646_{\,\pm\, .002}$& $.605_{\,\pm\, .002}$
    		& $.540_{\,\pm\, .001}$& $.621_{\,\pm\, .002}$& $.581_{\,\pm\, .001}$
\\
& BootEA	& \textcolor{cyan}{$.739_{\,\pm\, .014}$}& \textcolor{cyan}{$.849_{\,\pm\, .010}$}& \textcolor{cyan}{$.788_{\,\pm\, .012}$}
    		& \textcolor{red}{$.958_{\,\pm\, .001}$}& \textcolor{red}{$.984_{\,\pm\, .001}$}& \textcolor{red}{$.969_{\,\pm\, .001}$}
    		& \textcolor{cyan}{$.703_{\,\pm\, .004}$}& \textcolor{cyan}{$.827_{\,\pm\, .003}$}& \textcolor{cyan}{$.761_{\,\pm\, .003}$}
    		& \textcolor{blue}{$.886_{\,\pm\, .003}$}& \textcolor{blue}{$.944_{\,\pm\, .002}$}& \textcolor{blue}{$.912_{\,\pm\, .002}$}
\\
& GCNAlign	& $.465_{\,\pm\, .012}$& $.626_{\,\pm\, .011}$& $.536_{\,\pm\, .011}$
    			& $.875_{\,\pm\, .005}$& $.948_{\,\pm\, .004}$& $.907_{\,\pm\, .004}$
    			& $.528_{\,\pm\, .003}$& $.695_{\,\pm\, .005}$& $.605_{\,\pm\, .004}$
    			& $.620_{\,\pm\, .006}$& $.779_{\,\pm\, .006}$& $.693_{\,\pm\, .006}$
\\
& AttrE	& $.668_{\,\pm\, .012}$& $.803_{\,\pm\, .009}$& $.731_{\,\pm\, .010}$
    		& $.914_{\,\pm\, .015}$& $.97_{\,\pm\, .007}$& $.939_{\,\pm\, .012}$
    		& $.678_{\,\pm\, .017}$& $.81_{\,\pm\, .012}$& $.739_{\,\pm\, .015}$
    		& $.720_{\,\pm\, .01}$& $.846_{\,\pm\, .007}$& $.778_{\,\pm\, .008}$
\\
& IMUSE	& $.392_{\,\pm\, .013}$& $.571_{\,\pm\, .023}$& $.473_{\,\pm\, .017}$
    		& $.899_{\,\pm\, .011}$& $.949_{\,\pm\, .007}$& $.922_{\,\pm\, .009}$
    		& $.536_{\,\pm\, .018}$& $.700_{\,\pm\, .019}$& $.613_{\,\pm\, .018}$
    		& $.629_{\,\pm\, .011}$& $.774_{\,\pm\, .010}$& $.696_{\,\pm\, .010}$
\\
& SEA	& $.500_{\,\pm\, .011}$& $.706_{\,\pm\, .012}$& $.591_{\,\pm\, .012}$
    		& $.899_{\,\pm\, .005}$& $.950_{\,\pm\, .003}$& $.923_{\,\pm\, .004}$
    		& $.490_{\,\pm\, .042}$& $.677_{\,\pm\, .040}$& $.578_{\,\pm\, .040}$
    		& $.526_{\,\pm\, .021}$& $.687_{\,\pm\, .021}$& $.603_{\,\pm\, .021}$
\\
& RSN4EA	& $.514_{\,\pm\, .003}$& $.655_{\,\pm\, .004}$& $.580_{\,\pm\, .003}$
    			& \textcolor{cyan}{$.933_{\,\pm\, .003}$}& \textcolor{cyan}{$.974_{\,\pm\, .001}$}& \textcolor{blue}{$.951_{\,\pm\, .002}$}
    			& $.620_{\,\pm\, .002}$& $.769_{\,\pm\, .003}$& $.688_{\,\pm\, .002}$
    			& $.841_{\,\pm\, .003}$& \textcolor{cyan}{$.922_{\,\pm\, .001}$}& \textcolor{cyan}{$.877_{\,\pm\, .002}$}
\\
& MultiKE	& \textcolor{blue}{$.903_{\,\pm\, .004}$}& \textcolor{blue}{$.939_{\,\pm\, .003}$}& \textcolor{blue}{$.920_{\,\pm\, .003}$}
    		& $.856_{\,\pm\, .004}$& $.908_{\,\pm\, .002}$& $.881_{\,\pm\, .003}$
    		& \textcolor{blue}{$.884_{\,\pm\, .005}$}& \textcolor{blue}{$.920_{\,\pm\, .004}$}& \textcolor{blue}{$.901_{\,\pm\, .005}$}
    		& \textcolor{cyan}{$.853_{\,\pm\, .003}$}& $.896_{\,\pm\, .003}$& $.874_{\,\pm\, .003}$
\\
& RDGCN	& \textcolor{red}{$.931_{\,\pm\, .004}$}& \textcolor{red}{$.969_{\,\pm\, .003}$}& \textcolor{red}{$.949_{\,\pm\, .003}$}
    		& \textcolor{blue}{$.936_{\,\pm\, .003}$}& $.966_{\,\pm\, .001}$& \textcolor{cyan}{$.950_{\,\pm\, .002}$}
    		& \textcolor{red}{$.897_{\,\pm\, .001}$}& \textcolor{red}{$.950_{\,\pm\, .001}$}& \textcolor{red}{$.921_{\,\pm\, .001}$}
    		& \textcolor{red}{$.911_{\,\pm\, .002}$}& \textcolor{red}{$.949_{\,\pm\, .002}$}& \textcolor{red}{$.928_{\,\pm\, .002}$}
\\ \hline \multicolumn{14}{l}{$\textrm{Means}_{\,\pm\, \textrm{stds.}}$ are shown. Top-3 results on each dataset are marked in red, blue and cyan, respectively.  The same to the following.}
\end{tabular}

%% file: tab_csls.tex
\begin{tabular}{|l|cccc|}
	\hline \rowcolor{gray!10} & Greedy & Greedy w/ CSLS & SM & SM w/ CSLS\\ 
	\hline 
	MTransE & $.463_{\,\pm\, .013}$ & $.550_{\,\pm\, .009}$ & $.694_{\,\pm\, .006}$ & $.697_{\,\pm\, .010}$\\
	IPTransE & $.313_{\,\pm\, .009}$ & $.339_{\,\pm\, .013}$ & $.370_{\,\pm\, .018}$ & $.369_{\,\pm\, .018}$\\
	JAPE & $.469_{\,\pm\, .009}$ & $.549_{\,\pm\, .009}$ & $.692_{\,\pm\, .015}$ & $.691_{\,\pm\, .015}$\\
	KDCoE & $.661_{\,\pm\, .013}$ & $.679_{\,\pm\, .000}$ & $.840_{\,\pm\, .024}$ & $.815_{\,\pm\, .031}$\\
	BootEA & \textcolor{cyan}{$.739_{\,\pm\, .014}$} & $.741_{\,\pm\, .009}$ & $.783_{\,\pm\, .007}$ & $.782_{\,\pm\, .006}$\\
	GCNAlign & $.465_{\,\pm\, .012}$ & $.531_{\,\pm\, .008}$ & $.613_{\,\pm\, .008}$ & $.582_{\,\pm\, .010}$\\
	AttrE & $.668_{\,\pm\, .012}$ & \textcolor{cyan}{$.778_{\,\pm\, .012}$} & \textcolor{cyan}{$.845_{\,\pm\, .012}$} & \textcolor{cyan}{$.857_{\,\pm\, .012}$}\\
	IMUSE & $.392_{\,\pm\, .013}$ & $.448_{\,\pm\, .018}$ & $.520_{\,\pm\, .028}$ & $.518_{\,\pm\, .030}$\\
	SEA & $.500_{\,\pm\, .011}$ & $.557_{\,\pm\, .017}$ & $.647_{\,\pm\, .012}$ & $.650_{\,\pm\, .012}$\\ 
	RSN4EA & $.514_{\,\pm\, .003}$ & $.548_{\,\pm\, .003}$ & $.571_{\,\pm\, .002}$ & $.575_{\,\pm\, .004}$\\
	MultiKE & \textcolor{blue}{$.903_{\,\pm\, .004}$} & \textcolor{blue}{$.925_{\,\pm\, .003}$} & \textcolor{blue}{$.951_{\,\pm\, .003}$} & \textcolor{blue}{$.956_{\,\pm\, .002}$}\\ 
	RDGCN & \textcolor{red}{$.931_{\,\pm\, .004}$} & \textcolor{red}{$.956_{\,\pm\, .002}$} & \textcolor{red}{$.962_{\,\pm\, .002}$} & \textcolor{red}{$.979_{\,\pm\, .001}$}\\
	\hline
\end{tabular}

%% file: tab_conventional.tex
\begin{tabular}{|l|l|lll|lll|lll|lll|}
	\hline \rowcolor{gray!10} \multicolumn{2}{|c|}{} & \multicolumn{3}{c|}{15K (V1)} & \multicolumn{3}{c|}{15K (V2)} & \multicolumn{3}{c|}{100K (V1)} & \multicolumn{3}{c|}{100K (V2)} \\     
	\cline{3-14} \rowcolor{gray!10} \multicolumn{2}{|c|}{} & \multicolumn{1}{c}{Precision} & \multicolumn{1}{c}{Recall} & \multicolumn{1}{c|}{F1-score} & \multicolumn{1}{c}{Precision} & \multicolumn{1}{c}{Recall} & \multicolumn{1}{c|}{F1-score} & \multicolumn{1}{c}{Precision} & \multicolumn{1}{c}{Recall} & \multicolumn{1}{c|}{F1-score} & \multicolumn{1}{c}{Precision} & \multicolumn{1}{c}{Recall}& \multicolumn{1}{c|}{F1-score} \\ 
	\hline \parbox[t]{2mm}{\multirow{3}{*}{\rotatebox[origin=c]{90}{EN-FR}}}
	& LogMap & $.818_{\,\pm\, .002}$ & $.729_{\,\pm\, .002}$ & $.771_{\,\pm\, .001}$ & $.599_{\,\pm\, .003}$ & $.805_{\,\pm\, .001}$ & $.687_{\,\pm\, .002}$ & $.719_{\,\pm\, .001}$ & $.677_{\,\pm\, .001}$ & $.697_{\,\pm\, .001}$ & $.541_{\,\pm\, .001}$ & $.709_{\,\pm\, .000}$ & $.614_{\,\pm\, .001}$
	\\
	& PARIS & \textcolor{red}{$.907_{\,\pm\, .000}$} & \textcolor{red}{$.900_{\,\pm\, .000}$} & \textcolor{red}{$.903_{\,\pm\, .000}$} & \textcolor{red}{$.929_{\,\pm\, .001}$} & \textcolor{red}{$.939_{\,\pm\, .001}$} & \textcolor{red}{$.934_{\,\pm\, .001}$} & \textcolor{red}{$.852_{\,\pm\, .000}$} & \textcolor{red}{$.844_{\,\pm\, .000}$} & \textcolor{red}{$.848_{\,\pm\, .000}$} & \textcolor{red}{$.874_{\,\pm\, .000}$} & \textcolor{red}{$.888_{\,\pm\, .000}$} & \textcolor{red}{$.881_{\,\pm\, .000}$} \\ 
	& OpenEA & ${.755_{\,\pm\, .004}}^{*}$ & ${.755_{\,\pm\, .004}}^{*}$ & ${.755_{\,\pm\, .004}}^{*}$ & ${.864_{\,\pm\, .007}}^{\ddag}$ & ${.864_{\,\pm\, .007}}^{\ddag}$ & ${.864_{\,\pm\, .007}}^{\ddag}$ & ${.640_{\,\pm\, .004}}^{*}$ & ${.640_{\,\pm\, .004}}^{*}$ & ${.640_{\,\pm\, .004}}^{*}$ & ${.715_{\,\pm\, .003}}^{*}$ & ${.715_{\,\pm\, .003}}^{*}$ & ${.715_{\,\pm\, .003}}^{*}$
	\\
	\hline \parbox[t]{2mm}{\multirow{3}{*}{\rotatebox[origin=c]{90}{EN-DE}}}
	& LogMap & $.925_{\,\pm\, .002}$ & $.725_{\,\pm\, .001}$ & $.813_{\,\pm\, .001}$ & $.888_{\,\pm\, .000}$ & $.702_{\,\pm\, .002}$ & $.784_{\,\pm\, .001}$ & $.827_{\,\pm\, .001}$ & $.732_{\,\pm\, .001}$ & $.777_{\,\pm\, .001}$ & $.729_{\,\pm\, .003}$ & $.729_{\,\pm\, .001}$ & $.729_{\,\pm\, .002}$
	\\
	& PARIS & \textcolor{red}{$.938_{\,\pm\, .000}$} & \textcolor{red}{$.933_{\,\pm\, .000}$} & \textcolor{red}{$.935_{\,\pm\, .000}$} & \textcolor{red}{$.959_{\,\pm\, .000}$} & \textcolor{red}{$.964_{\,\pm\, .000}$} & \textcolor{red}{$.961_{\,\pm\, .000}$} & \textcolor{red}{$.890_{\,\pm\, .000}$} & \textcolor{red}{$.886_{\,\pm\, .000}$} & \textcolor{red}{$.888_{\,\pm\, .000}$} & \textcolor{red}{$.913_{\,\pm\, .000}$} & \textcolor{red}{$.923_{\,\pm\, .000}$} & \textcolor{red}{$.918_{\,\pm\, .000}$} \\
	& OpenEA & ${.830_{\,\pm\, .006}}^{*}$ & ${.830_{\,\pm\, .006}}^{*}$ & ${.830_{\,\pm\, .006}}^{*}$ & ${.833_{\,\pm\, .015}}^{\dag}$ & ${.833_{\,\pm\, .015}}^{\dag}$ & ${.833_{\,\pm\, .015}}^{\dag}$ & ${.722_{\,\pm\, .002}}^{*}$ & ${.722_{\,\pm\, .002}}^{*}$ & ${.722_{\,\pm\, .002}}^{*}$ & ${.766_{\,\pm\, .002}}^{*}$ & ${.766_{\,\pm\, .002}}^{*}$ & ${.766_{\,\pm\, .002}}^{*}$
	\\
	\hline \parbox[t]{2mm}{\multirow{3}{*}{\rotatebox[origin=c]{90}{D-W}}}
	& LogMap & \multicolumn{1}{c}{-} & \multicolumn{1}{c}{-} & \multicolumn{1}{c|}{-} & \multicolumn{1}{c}{-} & \multicolumn{1}{c}{-} & \multicolumn{1}{c|}{-} & \multicolumn{1}{c}{-} & \multicolumn{1}{c}{-} & \multicolumn{1}{c|}{-} & \multicolumn{1}{c}{-} & \multicolumn{1}{c}{-} & \multicolumn{1}{c|}{-} 
	\\
	& PARIS & \textcolor{red}{$.746_{\,\pm\, .001}$} & \textcolor{red}{$.723_{\,\pm\, .002}$} & \textcolor{red}{$.734_{\,\pm\, .001}$} & \textcolor{red}{$.828_{\,\pm\, .000}$} & \textcolor{red}{$.854_{\,\pm\, .001}$} & \textcolor{red}{$.840_{\,\pm\, .001}$} & \textcolor{red}{$.692_{\,\pm\, .000}$} & \textcolor{red}{$.643_{\,\pm\, .000}$} & \textcolor{red}{$.667_{\,\pm\, .000}$} & \textcolor{red}{$.783_{\,\pm\, .002}$} & \textcolor{red}{$.806_{\,\pm\, .002}$} & \textcolor{red}{$.795_{\,\pm\, .002}$} \\
	& OpenEA & ${.572_{\,\pm\, .008}}^{\dag}$ & ${.572_{\,\pm\, .008}}^{\dag}$ & ${.572_{\,\pm\, .008}}^{\dag}$ & ${.821_{\,\pm\, .004}}^{\dag}$ & ${.821_{\,\pm\, .004}}^{\dag}$ & ${.821_{\,\pm\, .004}}^{\dag}$ & ${.516_{\,\pm\, .006}}^{\dag}$ & ${.516_{\,\pm\, .006}}^{\dag}$ & ${.516_{\,\pm\, .006}}^{\dag}$ & ${.766_{\,\pm\, .007}}^{\dag}$ & ${.766_{\,\pm\, .007}}^{\dag}$ & ${.766_{\,\pm\, .007}}^{\dag}$
	\\
	\hline \parbox[t]{2mm}{\multirow{3}{*}{\rotatebox[origin=c]{90}{D-Y}}}
	& LogMap & \textcolor{red}{$.971_{\,\pm\, .001}$} & \textcolor{red}{$.943_{\,\pm\, .002}$} & \textcolor{red}{$.957_{\,\pm\, .001}$} & $.944_{\,\pm\, .002}$ & \textcolor{red}{$.984_{\,\pm\, .001}$} & $.964_{\,\pm\, .002}$ & \textcolor{red}{$.920_{\,\pm\, .001}$} & \textcolor{red}{$.915_{\,\pm\, .001}$} & \textcolor{red}{$.917_{\,\pm\, .001}$} & {$.954_{\,\pm\, .001}$} & $.912_{\,\pm\, .000}$ & $.933_{\,\pm\, .001}$
	\\
	& PARIS & $.899_{\,\pm\, .002}$ & $.869_{\,\pm\, .002}$ & $.884_{\,\pm\, .002}$ & \textcolor{red}{$.972_{\,\pm\, .000}$} & $.971_{\,\pm\, .000}$ & \textcolor{red}{$.971_{\,\pm\, .000}$} & $.893_{\,\pm\, .000}$ & $.866_{\,\pm\, .000}$ & $.880_{\,\pm\, .000}$ & \textcolor{red}{$.956_{\,\pm\, .000}$} & \textcolor{red}{$.953_{\,\pm\, .000}$} & \textcolor{red}{$.955_{\,\pm\, .000}$} \\
	& OpenEA & ${.931_{\,\pm\, .004}}^{*}$ & ${.931_{\,\pm\, .004}}^{*}$ & ${.931_{\,\pm\, .004}}^{*}$ & ${.958_{\,\pm\, .001}}^{\dag}$ & ${.958_{\,\pm\, .001}}^{\dag}$ & ${.958_{\,\pm\, .001}}^{\dag}$ & ${.897_{\,\pm\, .001}}^{*}$ & ${.897_{\,\pm\, .001}}^{*}$ & ${.897_{\,\pm\, .001}}^{*}$ & ${.911_{\,\pm\, .002}}^{*}$ & ${.911_{\,\pm\, .002}}^{*}$ & ${.911_{\,\pm\, .002}}^{*}$
	\\
	\hline
	\multicolumn{14}{l}{The results of RDGCN, BootEA and MultiKE are marked in $``*"$, $``\dag"$ and $``\ddag"$, respectively.}
\end{tabular}